\documentclass[12pt]{cmuthesis}

\usepackage{epigraph}

\usepackage{times}
\usepackage{fullpage}
\usepackage{graphicx}
\usepackage{amsthm,amssymb,amsmath}
\usepackage[numbers,sort]{natbib}
\usepackage[backref,pageanchor=true,plainpages=false, pdfpagelabels, bookmarks,bookmarksnumbered,
]{hyperref}
\usepackage{subfigure}
\usepackage[ruled, linesnumbered]{algorithm2e}
\usepackage[xcolor]{changebar}
\usepackage{dsfont}
\usepackage{multicol}

\usepackage[a4paper,twoside,vscale=.8,hscale=.75,nomarginpar, hmarginratio=1:1]{geometry}

\usepackage{mathtools}

\newcommand{\edited}[1]{%
    #1
  }%

\newcommand{\Var}{\mathrm{Var}}
\newcommand{\E}{\mathbb{E}}
\newcommand{\A}{\mathcal{A}}
\newcommand{\Sc}{\mathcal{S}}
\newcommand{\As}{\mathcal{A}_s}
\newcommand{\ot}{o_{1:t}}
\newcommand{\ok}{o_{1:k}}
\newcommand{\rt}{r_{1:t}}
\newcommand{\wt}{\omega_{1:t}}
\newcommand{\wk}{\omega_{1:k}} 
\newcommand{\lsa}{\lambda^{sa}_{\ot}}
\newcommand{\lsalpha}{\lambda^{s\alpha}_{\ot}}

\newcommand{\VOC}{\textsc{VOC}}
\newcommand{\VOCp}{\textsc{VOC$'$}}
\newcommand{\BSR}{\textsc{BSR}}
\newcommand{\BSRo}{\textsc{BSR}^*}
\newcommand{\SR}{\textsc{SR}}

\newcommand{\nsr}{{\langle s', a' \rangle \textrm{ is $n$ step reachable}}}

\newcommand*{\Scale}[2][4]{\scalebox{#1}{$#2$}}%

\theoremstyle{lemma}

\theoremstyle{theorem}

\newenvironment{hproof}{%
  \proof}{\endproof}

\theoremstyle{proposition}
\newtheorem{proposition}{Proposition}

\theoremstyle{remark}

\theoremstyle{definition}
\newtheorem{definition}{Definition}

\theoremstyle{corollary}

\theoremstyle{definition}
\newtheorem{example}{Example}

\makeatletter
\newenvironment{chapquote}[2][2em]
  {\setlength{\@tempdima}{#1}%
   \def\chapquote@author{#2}%
   \parshape 1 \@tempdima \dimexpr\textwidth-2\@tempdima\relax%
   \itshape}
  {\par\normalfont\hfill--\ \chapquote@author\hspace*{\@tempdima}\par\bigskip}
\makeatother

\begin {document} 
\frontmatter

\pagestyle{empty}

\title{ 
{\bf Computing the Value of Computation \\ for Planning}}
\author{Can Eren Sezener}
\date{October 2018}
\Year{2018}
\trnumber{}

\supervisor{
Prof. Peter Dayan \\ 
University College London
}





\maketitle
%

\pagestyle{empty} 

%
%
%
%
%
%
%
%
%
%
%

\begin{abstract}

An intelligent agent performs actions in order to achieve its goals. Such
actions can either be externally directed, such as opening a door, or
internally directed, such as writing data to a memory location or strengthening
a synaptic connection. Some internal actions, to which we refer as
computations, potentially help the agent choose better actions. Considering
that (external) actions and computations might draw upon the same resources,
such as time and energy, deciding when to act or compute, as well as what to
compute, are detrimental to the performance of an agent.

In an environment that provides rewards depending on an agent's behavior, an
action's value is typically defined as the sum of expected long-term rewards
succeeding the action (itself a complex quantity that depends on what the agent
goes on to do after the action in question). However, defining the value of a
computation is not as straightforward, as computations are only valuable in a
higher order way, through the alteration of actions.

This thesis offers a principled way of computing the value of a computation in
a planning setting formalized as a Markov decision process. We present two
different definitions of computation values: static and dynamic. They address
two extreme cases of the computation budget: affording calculation of zero or
infinitely many steps in the future. We show that these values have desirable
properties, such as temporal consistency and asymptotic convergence.

Furthermore, we propose methods for efficiently computing and approximating the
static and dynamic computation values. We describe a sense in which the
policies that greedily maximize these values can be optimal. Furthermore, we
utilize these principles to construct Monte Carlo tree search algorithms that
outperform most of the state-of-the-art in terms of finding higher quality
actions given the same simulation resources.

\end{abstract}

\begin{acknowledgments}
	
I feel incredibly privileged to have had Peter Dayan as my thesis supervisor. I
am especially appreciative of him always going through my work with great care
and providing helpful feedback. His mathematical rigor and scrutiny has greatly
helped me become a better thinker. Not every master's student is as lucky.

I am very thankful to Klaus Obermayer for being an excellent internal
supervisor and helping with many aspects of my thesis process and studies.

Special thanks to Mehdi Keramati for supervising my first lab rotation,
which has been pivotal for my research and career. In fact, the topics he has
introduced me to have strongly influenced my interests, and paved the way for
this thesis.

I am grateful to my bachelor's advisor Erhan {\"O}ztop for enabling me to
experience the joy of scientific discovery and consequently setting me on the
path of research.

My master's journey has been a long one. I would like to thank my friends and
colleagues for making it also an enjoyable one: Murat and H{\"u}seyin---for
being excellent general-purpose buddies, Milena---for the
thesis-writing-solidarity, Maraike---for the sanity-boosting, the Gatsby
Unit---for providing an excellent, high information-content environment during
my six month visit, and the BCCN family, especially my classmates, ``the year
above''-mates, the faculty, Robert, and Margret---for many things, enumeration
of which could only be done justly in the Supporting Information.

Many thanks to colleagues whose input has shaped this thesis: Florian Wenzel
for giving me a be a better grasp of Gaussian processes, Georg Chechelnizki for
applying his logic superpowers to crush the trivial xor inconsistent statements
on the abstract and helping with the writing of ``Zusammenfasssung'', Greg
Knoll for his feedback on my thesis defense, and Fiona Mild for all the
extracellular dopamine.

Additional thanks to TEV and DAAD for the generous scholarship that made my
studies significantly more comfortable.

Finally, I am incredibly grateful to my parents, \c{C}i\u{g}dem and Erdal, for
their unconditional support.

\end{acknowledgments}

\tableofcontents

\mainmatter


%
%
%
%
%


\pagestyle{plain}
\chapter{Introduction}
\begin{chapquote}{Epictetus, \textit{Discourses}}
``First say to yourself what you would be; and then do what you have to do.''
\end{chapquote}

Imagine you are playing chess in a tournament, contemplating your next move.
You have already considered some sequences of moves, and it seems like moving
your c-pawn is your best bet---but you are not entirely sure. Should you think
further? Or should you save your time for later in the game? Let us say you
decide that you are at a critical point in game, and thinking further is well
worth your limited clock-time. Then should you spend your time considering the
c-pawn move further, to see if it as good as you think it is? Or should you
instead think about bringing your bishop into the game, which seems like your
best alternative? Maybe, instead, you should look around on the board to see if
there are any promising moves that you have not considered yet?

Such questions do not only concern chess players. Rather, variants of them
present themselves on a daily basis, where one has to decide between thinking
and acting, as well as what to think about. Reckless behavior,
``analysis-paralysis'', or rumination might be seen as negative consequences
resulting from too little, or too much, thinking than is ideal; or thinking
about the ``wrong'' things.

An intelligent robot (or its programmer) has to deal with the same questions as
well. For a mobile robot, acting and computing (i.e., thinking) often draw upon
the same resources such as time and energy. Therefore, a resource-bounded robot
needs a policy to decide when to act, when to compute, and what to compute.

In this thesis, we address these questions by proposing ways in which one could
assign values to computations---which is typically\footnote{We build upon the
existing body of literature on VOC-like measures, which we discuss later in
this chapter.} referred to as the \emph{value of computation} (VOC). We do so
by operating in a Bayesian framework, where the agent is uncertain about the
values of possible actions, and decreases its uncertainty by performing
computations. 

As we later discuss, this is not a straightforward task for a variety of
reasons. For instance, the value of an action does not only depend on the
present computation, but it also depends on future computations. A chess player
running out of clock-time might decide to ignore a seemingly complicated move
if she thinks she might not have enough time to think properly through all the
complications resulting from the move.

This problem could be addressed by our proposal to some extent. We propose
static and dynamic action value functions, which respectively cover the the two
extremes of zero and infinitely many future computations. Furthermore, we argue
that the current work on VOC-like measures and relating topics are limited in
two severe ways: (I) they exclusively focus on what we refer to as static
values---ignoring effects of future computations, (II) they assume a
bandit-like scenario where action values are stationary---whereas, in planning,
action values improve in expectation as the agent performs more computations.
Our formulations address these two limitations.

In the theoretical part of the thesis, we discuss the properties of static and
dynamic values, and define the value of computation as the expected increase in
these values. We show that choosing static/dynamic-value-maximizing
computations is optimal in certain cases. Furthermore, we provide exact and
efficient approximate algorithms for computing these values. In the applied
part of the thesis, we give a Monte Carlo tree search algorithm based on
VOC-maximization. We empirically demonstrate its efficacy by comparing it
against other popular or similar methods in two different environments.


Before moving on to our discussion of computation values in the next chapter,
we first provide some mathematical background, which will form the basis of our work. Then we discuss some of the work relating to this thesis below. Lastly provide an outline of the thesis.

\section{Preliminaries}
Below, we introduce multi-armed bandit problems and Markov decision process, which we will utilize heavily throughout this thesis.

\subsection{Multi-armed bandit problems}

A multi-armed bandit (MAB, or simply \emph{bandit}) problem is composed of a
set of actions $\A$, where each action is associated with a distinct reward
distribution function. Let $Q : \A \rightarrow \mathbb{R}$ be the expected
reward function: $Q(a) = \E[ r_t | a_t = a]$.

The goal then is to minimize some form of \emph{regret}. In this thesis, we
define it as the expected reward one misses out by taking action $a_t$, $R(t) =
\max_a Q(a) - Q(a_t)$. For instance, one can aim to minimize \emph{cumulative
regret}, $\sum_{t=0}^\infty R(t)$; or simply the regret $R(t)$ at a particular
$t$, which is referred to as \emph{simple regret}. In either case, the actor
cannot directly observe the regret, as $Q$ is unknown, but typically forms
estimates of it.

\subsection{Markov decision processes}
A finite Markov decision process (MDP) is a $5$-tuple $\langle \Sc, \A, \mathcal{P}, \mathcal{R}, \gamma \rangle$, where
\begin{itemize}
\item $\Sc$ is a finite set of states
\item $\A$ is a finite set of actions
\item $\mathcal{P}$ is the transition function such that
  $\mathcal{P}^a_{ss'} = P(s_t = s' | s_t = s, a_t = a)$, where $s,s' \in
  \Sc$ and $a \in \A$
\item $\mathcal{R}$ is the expected immediate reward function such
  that $\mathcal{R}^a_{ss'} = \E[r_{t+1} | s_t = s, a_t = a, s_{t+1} =
  s']$, where again $s,s' \in \Sc$ and $a \in \A$
\item $\gamma$ is the discount factor such that $\gamma \in [0, 1)$.
\end{itemize}

We assume an agent interacts with the environment via a (potentially
stochastic) policy $\pi$, such that $\pi(s,a) = P(a_t = a | s_t = s)$. This
quantity is typically conditioned on some parameters which determine the policy
of the agent, which is left out from the notation here. The agent aims to
maximize the the expected value of the cumulative discounted rewards
$\mathbb{E}_\pi \left[ R_t \right | s_t = s]$, where $R_t = \sum_{i=0}^{\infty}
{\gamma}^i r_{t+i}$, by controlling its policy $\pi$. 
 
These expected values are typically estimated by planning and reinforcement
learning methods in some form. As such, they get a preferential treatment in
terms of the notation, and simply get referred to as \emph{values}, which is
given by \emph{value functions}. The value of a state $s$ is given by the value
function $V^\pi$:
\begin{equation}
V^\pi(s) = \mathbb{E}_\pi \left[ R_t \right | s_t = s] \: .
\end{equation}
Similarly, the value of a state-action $s, a$ is given by the value function $Q^\pi$:
\begin{equation}
Q^\pi(s, a) = \mathbb{E}_\pi \left[ R_t \right | s_t = s, a_t=a] \: .
\end{equation}

Both of these value functions can be defined recursively, by relating a state (or a state-action) to its neighboring states (or state-actions), by the \emph{Bellman equations}:
\begin{align}
Q^\pi(s,a) &= \sum_{s'}\mathcal{P}_{ss'}^{a} \left[ \mathcal{R}_{ss'}^{a} +  \gamma \sum_{a'} \pi(s',a') Q^\pi(s', a') \right] \\
V^\pi(s) &= \sum_a \pi(s,a) \sum_{s'}\mathcal{P}_{ss'}^{a} \left[ \mathcal{R}_{ss'}^{a} +  \gamma V^\pi(s') \right] \: .
\end{align}

Solving a planning or a reinforcement learning problem boils down to
find an optimal policy $\pi^*$ that is better than all other policies. 
In other words, for an optimal policy $\pi^*$ 
\begin{equation}
\pi^* = \arg \max_{\pi} V^\pi(s)
\end{equation}
must hold for all $s$.

State and state-action values of an optimal policy are given
respectively by the \emph{optimal state value function} $V^*$ and the
\emph{optimal state-action value function}
$Q^*$. Formally,
\begin{equation}
V^*(s) = V^{\pi^*}(s) = \max_\pi V^\pi(s)
\end{equation}
for all $s$ and
\begin{equation}
Q^*(s,a) = Q^{\pi^*}(s,a) = \max_\pi Q^\pi(s,a) \: ,
\end{equation}
for all $s$ and $a$.
These optimal value functions can also be defined recursively, through what we refer to as the \emph{Bellman optimality equations}:
\begin{align}
Q^*(s,a) &= \sum_{s'}\mathcal{P}_{ss'}^{a} \left[ \mathcal{R}_{ss'}^{a} +  \gamma \sum_{a'} \max_{a'} Q^*(s', a') \right] \\
V^*(s) &= \max_{a'} \sum_{s'}\mathcal{P}_{ss'}^{a} \left[
  \mathcal{R}_{ss'}^{a} +  \gamma V^*(s') \right] \: . 
\end{align}

\section{Related work}
We draw upon two different lines of research. The first considers the value of
resolving one's uncertainty, and relates to ideas such as the value of
information, knowledge gradients, and metareasoning. The second line is
concerned with finding approximately optimal actions in known
environments---that is, approximate planning.

\subsection{Information values and metareasoning}

We draw heavily on Howard's work on \emph{information value theory}
\cite{howard1966}, which is concerned with the economic advantage gained
through reducing one's uncertainties. The original form of this work addresses
the questions of how much one should pay to obtain a piece of information. The
problem we investigate in this thesis is analogous. However, instead of the
value of an external piece of information, we are interested in the value of an
internal piece of information, which results from thinking or computing.

Formal approaches to this problem include the work of Russell and Wefald
\cite{wefald1991}. This lays down the theoretical foundations of \emph{rational
metareasoning} which concerns finding and performing optimal internal actions
(i.e., computations). However, as noted by the authors, solving the
metareasoning problem optimally---meaning performing the optimal sequence of
computations---is intractable for several reasons. One of the reasons is that
optimal meteareasoning is a harder computational problem than optimal
reasoning, as recently proven by Lin et. al. \cite{lin2015} for a general class
of problems. This is because, finding the optimal solution to the metareasoning
problem comes with a great computational burden at the level of
meta-metareasoning, optimization of which requires even a more meta reasoner.
In other words, optimality at one level can only be achieved with a great cost
at the one higher (more meta) level. One reason for this is that (meta) state
and action spaces grow as one goes up the metareasoning hierarchy. In many
settings with a good reasoner, perfect metareasoning is likely to perform worse
than no-metareasoning---in the sense that the latter obtains the same solution
but at a lower computational cost. This being said, \emph{some} metareasoning
can be better than the two extremes.

We can perhaps better understand this with a neural networks analogy. Consider
three ways of updating the weights of a network: random walk, stochastic
gradient descent (SGD), and Newton's method. Random walk is the cheapest to
compute (similar to no-metareasoning) but performs the worst. Newton's method
is similar to perfect meta-reasoning as it is the costliest to compute, but is
locally optimal. However, SGD, which falls between the two in terms of
computational costs performs often the best---even though its updates are noisy
and thus suboptimal. One reason is that the computations spent on performing
one ``optimal'' update can be more spent on performing many sub-optimal
updates, which often yields better results in practice.

Thus, Hay and others \cite{hay2012, hay2016} utilize ``myopic''
metareasoning---which is akin to performing computations that maximize the
immediate (rather than the long-term) rewards of the meta-level
problem\footnote{The definition of which can be found in the cited work.}---to
outperform non-meta reasoners in the context of approximate planning.

A parallel line of progress has recently been made by Ryzhov, Frazier, Powell,
and their colleagues \cite{frazier2009a, ryzhov2012, frazier2017}. They propose
the \emph{knowledge gradient (KG)}---which closely resembles Howard's
information values---and show that it solves Bayesian optimization
(specifically, selection and ranking) problems effectively.

We can consider two kinds of planning problems: ``stateless'' and ``stateful''.
In the former, outcomes of actions are path independent. That is, if two action
sequences are permutations of each other, then they must be equally rewarding
in expectation. Whereas, path-dependence may exist in the latter kind. To
elaborate, in stateful problems, actions not only yield rewards, but they also
modify the state of the agent. Therefore, the ordering of actions do impact the
reward expectations. Multi-armed bandits are examples of stateless problems,
and Markov decision processes of stateful problems typically.

To the best of our knowledge, all the work on rational metareasoning and
knowledge gradients either focus on stateless problems, or they treat stateful
problems as stateless problems. For example, in rational-metareasoning-based
Monte Carlo tree search \cite{hay2012}, the authors make a simplifying
statelessness assumption by treating the root actions as ``bandit arms''
with fixed quantities of expected rewards attached---even though the rewards
depend on the policy of the agent, and thus are not fixed. As also noted by the
authors, this is perhaps the most significant shortcoming of their method. In
this thesis we partially lift this constraint, by extending metareasoning (or
knowledge gradients) to distal (i.e., non-immediate) actions. Our remedy is
only partial, because the extent of actions for which metareasoning is to be
leveraged should be determined in advance. ``Stateful'' problems are typically
referred to as planning problems, which we explain next.

\subsection{Approximate planning}

Planning consists of finding an optimal sequence of external actions given an
environment and a goal. When formalized as a MDP, the goal is then to find an
optimal or a near optimal policy. However, achieving optimality is typically
intractable; and one has to resort to approximately optimal solutions. Monte
Carlo tree search (MCTS) is a popular family of methods for finding such
solutions. MCTS methods build a search tree of possible future states and
actions by using a model to simulate random trajectories. Good MCTS methods
exploit nascent versions of the tree to make judicious choices of what to
simulate. One main computational advantage of MCTS comes from the concept of
\emph{sparse sampling} \cite{kearns2002}. It turns out that it suffices to
build a backup (expectimax) tree that is sparse, and as such, to avoid the
curse of dimensionality associated with building complete backup trees.

UCT \cite{kocsis2006}, which we describe in detail later, is a popular MCTS
method that performs very well in many domains. Roughly speaking, UCT breaks
down how simulations are chosen in MCTS into smaller sub-problems, each of
which is then treated as a bandit problem. UCT is asymptotically
optimal---meaning, given infinite computation time, finds the best action.
However, it is also known to be biased in the sense that it explores less than
is optimal, when the goal is to minimize simple regret as in MCTS
\cite{bubeck2011}.

This is because UCT treats MCTS---where the goal is to minimize the (simple)
regret of the action eventually taken---as a traditional bandit problem---where
the goal is to minimize the \emph{cumulative} regret. To put it in another way,
in MCTS, simulations are valuable, only because they help select the best
action. However, UCT actually aims to maximize the sum of rewards obtained in
\emph{simulations}, rather than paying attention to the quality of actual
(i.e., not simulated) actions. Consequently,
it tries to avoid simulations with potentially low rewards, even though they
might help select better actions. In fact, Hay et. al. \cite{hay2012} improve
the performance of UCT just by adding a very limited metareasoning capability,
namely by computing an information value score for picking the immediate action
from which to run a simulation, and then resorting to vanilla UCT for the rest
of the decisions down the tree. The practical part of this thesis is concerned
with pushing this envelope by building MCTS methods with better metareasoning
capabilities.

\section{Thesis organization}
The thesis is organized as follows. In Chapter 2, we introduce different
notions of computation values and provide formal definitions. In Chapter 3, we
discuss how computation values can be computed or approximated efficiently. In
Chapter 4, we materialize our ideas into a Monte Carlo tree search algorithm
which leverages computation values. In Chapter 5, we evaluate our methods,
together with other popular methods, in two different environments, and show
that our computation-value-based algorithms perform either better than all
other alternatives, or comparably to the best alternative.

\chapter{Value of Computation}
\begin{chapquote}{Michel de Montaigne, \textit{Essays}}
``In truth, either reason is a mockery, or it must aim solely at our contentment, and the sum of its labors must tend to make us live well and at our ease...''
\end{chapquote}

We begin by giving a general formal definition of computations and computation
value. Then, we introduce additional assumptions, and consider multi-armed
bandit problems and Markov decision processes.

\section{Computations as higher-order functions}

Let us leave MDPs and bandits aside for a moment, and consider a more abstract setting.
Let $\Sc$ be the set of possible states of the agent, $\A$ be
the set of possible actions, and $Q: \Sc \times \A \rightarrow \mathbb{R}$ be the
value function of the agent indicating the desirability of state-actions.
\edited{Generally, we
consider an action desirable, if it is expected to lead to a valuable outcome.}

Assume the agent has some control over $Q$---\edited{which can either result from
obtaining better knowledge of the possible outcomes of actions and
their respective probabilities,}or it can be a
more direct form of control, like deeming hard-to-reach-grapes as sour as in
Aesop's fable. The
latter corresponds to altering of utility functions\footnote{Which assigns values to \emph{outcomes.}} (or primary reinforcers);
\edited{and the former of outcome probabilities, and thus of expected utilities} (or secondary
reinforcers) as well.  We refer to both kinds of alterations as
computations, and to both utilities and expected utilities as values for now.
When we are discussing computation values in Markov decision processes and
multi-armed bandits, we will provide more concrete definitions, and limit the
effect of computations to expected utility alterations only.

\edited{
In real time planning schemes such as Monte Carlo tree search (MCTS)---which we introduce in a subsequent chapter---the agent aims to maximize
the  value\footnote{In MCTS, there are \emph{true} values (i.e., those given by the Bellman optimality equation) which are determined by the environment. However, at the moment, we are considering a more abstract and general notion of values.}
of the best action, $\max_{a \in \As} Q(s, a)$, where $\As \subseteq \A$ is the set of actions available at state $s$, by performing computations. In
general, we can think of a computation as a higher-order function transforming
the value function, $\omega: (\Sc \times \A \rightarrow \mathbb{R}) \rightarrow
(\Sc \times \A  \rightarrow \mathbb{R})$. The value of the best action at state $s$ after
performing the computation will be $\max_{a \in \As} \omega[Q](s, a)$, causing a net
change of
\begin{equation}
\max_{a \in \As} \omega[Q](s, a) - \max_{a \in \As} Q(s,a) \: .
\end{equation} 
This is the value of a \emph{deterministic} computation whose outcome is known to the agent, which the agent aims to
maximize via,
\begin{equation}
\arg \max_{\omega \in \Omega} \left[ \max_{a \in \As} \omega[Q](s,a) - \max_{a \in \As} Q(s,a) \right] \: ,
\end{equation} 
where $\Omega$ is the set of possible computations.

Even though we would like to determine the optimizing computation so that we
could perform it, how to do it in an advantageous way is not clear. If one has
to actually perform all the computations in order to pick the best one, it is
likely to be counterproductive from a cost-benefit point of view as it would
require paying the cost of $|\Omega|$ many computations in return for
performing a single computation.}

However, if the agent has predictions concerning the effects of computations,
possibly obtained by storing the descriptive statistics of previous
computations' outcomes, this problem can be circumvented. We operate under this
assumption, and redefine $\omega$ as a random-valued higher-order function that maps a value function to a random
value function---\edited{as stochastic computations include deterministic ones as a subset.}
Then, $\omega[Q](s,a)$ for a known $a$
is a random variable in $\mathbb{R}$. This reflects the fact that the agent is
not entirely sure about the effect of a computation, and instead has a
probability measure over the set of possible effects.
Then, we obtain the value of computation ($\VOC$) for $\omega$
as
\begin{equation}
\VOC(\omega) = \E[\max_{a \in \As} \omega[Q](s,a)] - \max_{a \in \As} Q(s,a) \: .
\label{VOC}
\end{equation} 
On the other hand, the commonly used definition \cite{wefald1991, hay2012, tolpin2012} is slightly different, which we denote with $\VOCp$,
\begin{equation}
\VOCp(\omega) = \E[\max_{a \in \As} \omega[Q](s,a)] - \E[Q(\arg \max_{a \in \As}\omega[Q](s,a))] \: .
\label{VOC'}
\end{equation}
Let us rewrite the $\VOC$ definition in Equation~\ref{VOC} to make it more compatible with Equation~\ref{VOC'},
\begin{align}
\VOC(\omega) &= \E[\max_{a \in \As} \omega[Q](s,a)] - \E[Q(\arg \max_{a \in \As}Q(s,a))] \label{eq:vocomega} \\
&= \E[\max_{a \in \As} \omega[Q](s,a)] - Q(\arg \max_{a \in \As}Q(s,a)) \: .
\end{align}

The difference is subtle. $\VOCp$ uses the \emph{new} expected value of the
previously best action as the baseline; whereas $\VOC$ uses the old value of
the previously best action. Thus, $\VOCp$ only takes the differences due to
policy changes into account. Consider a case where there are objective values
to actions, meaning they are solely given by the environment such as in a
multi-armed bandit task, which we will formalize later on. In this case,
computations cannot directly modify the values ($Q$); however, they are useful
because they can modify the agent's posterior expectations of $Q$, which in
turn benefits the agent because he can make a better decision concerning what
action to take. $\VOCp$ assumes this setting of objective $Q$'s and therefore
values a computation solely based on its impact on his policy \emph{at his
current state}, i.e. $\pi(s, \cdot)$.

Lastly, it follows from the definition in Eq~\ref{VOC'} that $\forall
\omega: \VOCp(\omega) \ge 0$. However, we cannot say much about $\VOC$ yet. For
this, we need further assumptions, which we introduce next.

\section{Computations as evidence}

\edited{Up until now, we have assumed $Q$, the value function, is known to the agent.
Here, we relax this assumption, and treat it as a quantity that the agent estimates by performing computations.
Computations might achieve this in
two distinct ways: by resolving \emph{empirical} uncertainty or
\emph{computational} uncertainty. The former uncertainty is due to insufficient
external data. The latter type is due to insufficient processing of the
existing external data. Consider the problem of estimating the parameters of a
generative model from a dataset using Markov Chain Monte Carlo. In order to get
a more accurate estimate, one can either collect more data, or run more Monte
Carlo simulations. Alternatively, consider a tennis robot getting ready for a
shot. He would like to reduce his uncertainty about what the best motor
commands are. He can achieve this by collecting more data concerning the
velocity and the position of the ball using his sensors. Or, he can use the
data he has, but spend processor cycles unrolling the physics model he has into
the future. In many cases, the total uncertainty is a combination of both
empirical and computational uncertainties. That being said, we focus on
resolving uncertainties of this latter kind---that is, computational---and thus
consider planning in fully known environments. However, the results we obtain
should transfer to the cases where the agent also has to learn the statistics
of the environment (e.g., as in model-based learning).}

Agents that operate in unknown environments typically collect statistics
concerning the outcomes of their actions in order to make better decision. We
take this a step further, and assume the agent also keeps statistics concerning
his past computations and their outcomes. 
\edited{Subsequently, we take a Bayesian approach and assume the value function,
$Q: \mathcal{S} \times \mathcal{A} \rightarrow \mathbb{R}$, is sampled from a
known distribution, which could be given by the past experience of the agent.
\edited{Note that, as before, we do not enforce the Bellman equation constraint
yet, and treat the $Q$-function in a more general sense---state-actions that
have higher $Q$-values are more desirable.} The agent estimates this unknown
$Q$-function by performing computations, which transform $Q$, which is a random
valued function, to another random valued function. Here, we change our
formalization of computations for one last time. A computation $\omega$ is no
longer a higher-order function, but is a function that outputs imperfects
estimates of state-action values, which we refer to as observations. At a given
time, the agent will have performed a sequence of computations $\wt = \omega_1,
\omega_2, \dots \omega_t$ whose respective outcomes are $\ot = o_1, o_2, \dots,
o_t$. Each computation belongs to a finite set $\Omega$, and each outcome
belongs to a potentially infinite set $\mathcal{O}$. Thus, $\wt \in \Omega^t$
and $\ot \in \mathcal{O}^t$. For notational simplicity, we assume that the
outcome of a computation formally includes information about the computation
from which the outcome resulted. That is, if a computation $\omega$ yields
information about an $Q(s,a)\in \mathbb{R}$ then the outcome $o$ of $\omega$ is
in $\Sc \times \A \times \mathbb{R}$---as it carries information about a state
action, and its value\footnote{More generally, a computation might carry
information about multiple $\langle s,a \rangle$, then the outcome of the
computation would be in $(\Sc \times \A \times \mathbb{R})^n$.}}. The agent uses
$\ot$ to get a better estimate of $Q$, namely the posterior distribution $p(Q
|\ot)$. We introduce a random function $Z$, whose domain is $\mathcal{S} \times
\mathcal{A} \times \mathcal{O}^t$, such that $p(Z(s,a | \ot)) = p(Q(s,a) |
\ot)$ for all state-actions, or equivalently $Z(s,a | \ot) = Q(s,a) | \ot$. In other words,
$Z$-function is the posterior distribution of the unknown
$Q$-function conditioned on $\ot$.

We would like to evaluate the value of a computation prior to performing the
computation itself. Therefore, we can think of a meta-computation, which
evaluates the potential outcomes of a computation $\omega$. More specifically, we assume
computations draw samples from $Q(s,a) + \epsilon$ where $\epsilon$ is a
stochastic noise term capturing the imperfection of computations; and thus, meta-computations draw
samples from $Z(s,a | \ot) + \epsilon$, which is the posterior predictive
distribution. An alternative formulation for meta-computations would be to draw
samples directly from $Z(s,a | \ot)$, without including the noise term
$\epsilon$. This, in a sense, captures how much one can learn by performing
sufficiently many computations such
that the noise cancels out due to the Law of Large Numbers, which is equivalent
to the \emph{expected improvement} formulation. However, we would like to
quantify how much we could learn from a single computation; thus, the noise is
included.

\edited{
Previously, when computations were higher-order functions, we had denoted the
expected value of a function $f$, after a computation $\omega$ as
$\E[\omega[f](\cdot)]$. Now, $\omega$ yields an observation $o$ about $f$, rather than
transforming it. Therefore, we now denote the same value with $\E_{o \sim
\omega}[f(o)]$, which captures the expected value of $f$ after performing
$\omega$. The process of evaluating this expression is a meta-computation as it
carries information about computation $\omega$.} In most settings we will
introduce, the cost of performing a meta-computation will be negligible
compared to the cost of performing a computation. As such, we only apply the
cost-benefit analysis to the computations themselves.

We assume there is a one-to-one mapping between actions and computations for a
given state. That is, between $\mathcal{A}_s$, actions available at $s$, and
$\Omega_s$, computations that directly affect $Q(s, \cdot)$, such that each
$\omega \in \Omega_s$ corresponds to observing $Q(s,a)$ for a specific $a \in
\mathcal{A}_s$ and vice versa.
Therefore, from now on, performing a computation and sampling an action is used
interchangeably. This assumption is not necessary but convenient to introduce
now as it is both intuitive and common in planning algorithms. Consider the UCT
algorithm \cite{kocsis2006} as an example, for the computation called a
roll-out, whose goal is to provide information about the value of root actions.
 Note that, despite this one-to-one mapping between
computations and actions, a computation can have an effect on multiple
state-action values given the potential dependencies. The hypothetical outcome
$o \sim \omega$ of a candidate computation $\omega$, affects $Z(s,a | \ot)$ for
all $\langle s,a \rangle$ and transforms it to $Z(s,a | \ot \oplus o)$ where
$\oplus$ is the concatenation operator.

We avoid using the notation $\bar{o}_{t+1} $ for $\ot \oplus o$, because $\ot$
includes the outcomes of the previously performed computations, whereas $o$ is
the hypothetical outcome of a candidate computation, which has a different
distribution (e.g., posterior predictive) than those computations in $\ot$.

However, both past computations and future computations have the same type,
i.e., they belong to $\mathcal{O}$. As such, they can form a sequence when
concatenated. \edited{Note that, for a given $o$, $\E[Z(\cdot,\cdot | \ot \oplus
o)]$ is a
value in $\mathbb{R}^{|\Sc| \times |\A|}$}. However, for an unknown $o$, it is a random variable,
and $Z(\cdot, \cdot | \ot \oplus o)$ is a random function. 

For all $\langle s,a \rangle$, we have
\begin{equation} \E_{o \sim \omega}\left[ \E[Z(s,a | \ot \oplus o)] \right] =
\E[Z(s,a | \ot)]\: . \label{tower2} 
\end{equation} 

due to the law of total
expectation. This equality suggests that the beliefs of the agent are coherent
and cannot be exploited by a Dutch book. In our formulation, this is achieved
trivially, because we assume we know the prior distribution of $Q$. \edited{In the
experiments section we show that, even when we do not know the true sampling distribution of $Q$ (i.e., when we assume one),
algorithms based on this principle can perform very well.} 
Furthermore, we
postulate that humans do likely leverage some form informed priors, as we do
here, which can be satisfied by a TD-learning-like rule that
learns the second moments of action values in addition to the first ones.

If we consider the maximum of action values under computations, given $\max$ is a convex function, we have 
\begin{equation}
\E_{o \sim \omega}\left[ \max_{a \in \mathcal{A}_s}\E[Z(s,a | \ot \oplus o)] \right] \ge \max_{a \in \mathcal{A}_s}\E[Z(s,a | \ot)]\: ,
\label{jensen}
\end{equation}
due to Jensen's inequality. Most results in this chapter are derived
directly from these two relations. Equation~\ref{tower2} states that although
the posterior distribution of $Q$ may change after a computation, its
\emph{expectation} remains the same. Despite this, the maximum posterior
expectation is expected to increase or remain constant, as implied by
Equation~\ref{jensen}. In this setting, let us reintroduce the value of
computation definitions which we will use for the rest of
the text: 

\edited{
\begin{definition} We (re)define the value of computation,
\begin{equation}
\VOC(\omega ; s, \ot) = \mathbb{E}_{o \sim \omega}\left[\max_{a \in \mathcal{A}_s} \mathbb{E} [Z(s,a |  \ot \oplus o)] \right] - \max_{a \in \mathcal{A}_s} \mathbb{E}_{o \sim \omega} [Z(s,a | \ot) ] \label{voc2}\: ,
\end{equation}
where $o \sim Z(s,a | \ot) + \epsilon$ for some $\langle s,a \rangle$ and a noise term $\epsilon$.
\end{definition}
}

\begin{definition} The alternative different formulation from Eq~\ref{VOC'} is
\begin{align} \VOCp(\omega ; s, \ot) &= \mathbb{E}_{o
\sim \omega}\left[\max_{a \in \mathcal{A}_s} \mathbb{E} [Z(s,a | \ot \oplus o)]
\right. \nonumber \\ 
& \left. - \mathbb{E} \left[ Z\left(s, \arg \max_{a \in
\mathcal{A}_s} \mathbb{E}\left[ Z(s,a | \ot) \right]) | \ot \oplus o \right)
\right] \right] \: . 
\label{vocp} 
\end{align} 
\end{definition} 
We will compare these two definitions through out this section. Recall that for
now we assume computations directly inform us about the actions available at
the current state $s$. More specifically, a computation draws samples from a
specific action at $s$ and is used to potentially update value estimates of all
state-actions. Then we have:
\begin{proposition} If $\omega \in \Omega_s$ where $s$ is the current state of 
the agent, then
\begin{equation}
\VOC(\omega; s, \ot) = \VOCp(\omega; s, \ot)\: .
\end{equation}
\label{eq_voc}
\end{proposition}

\begin{proof} We need to show the second terms of Equations~\ref{voc2} and 
\ref{vocp} are equal.  Note that
\begin{equation}
\max_{a \in \mathcal{A}_s} \mathbb{E} [Z(s,a | \ot) ] = \mathbb{E} \left[ Z\left(s, \arg \max_{a \in \mathcal{A}_s}  \mathbb{E}\left[ Z(s,a | \ot) \right]) | \ot \right) \right]
\end{equation}
by rearrangement. Then we can utilize the law of total expectation to
assert,
\begin{equation}
\mathbb{E} \left[ Z\left(s, \arg \max_{a \in \mathcal{A}_s}  \mathbb{E}\left[ Z(s,a | \ot) \right]) | \ot \right) \right] = \mathbb{E} \left[ Z\left(s, \arg \max_{a \in \mathcal{A}_s}  \mathbb{E}\left[ Z(s,a | \ot) \right]) | \ot \oplus o \right) \right] \: .
\label{total}
\end{equation}
\end{proof}

A policy decides which action to take given a state. We now introduce
\textbf{meta-policies}, which decide what computation to perform given a state
and a sequence of past computation outcomes. A more formal treatise of
meta-policies and meta-MDPs can be found in \cite{hay2012}.

\begin{definition} \textbf{VOC-greedy} and \textbf{VOC$'$-greedy} are
meta-policies that perform the computation $\omega$ that maximizes $\VOC$ and
$\VOCp$ respectively until a fixed number of computations
is reached or until the $\VOC$ becomes non-positive for all computations. Having performed $t$
many computations whose outcomes are $\ot$, if the agent takes an action, then
he performs $\arg \max_{a \in \mathcal{A}_s} \mathbb{E} \left[Z(s,a | \ot)
\right]$ 
\end{definition}

Instead of performing a fixed number of computations, the agent might want to
decide whether the computation is worthwhile, and only if so, perform it. An
intuitive and somewhat \cite{hay2012, wefald1991} popular stopping criterion is
\begin{equation}
\max_{\omega \in \Omega} \VOC(\omega; s, \ot) < c \:,
\label{eq:stop}
\end{equation}
where $c$ is the cost of performing a computation, typically in terms of time
or energy.

We would like to evaluate the performance of meta-policies. The error
function we are interested minimizing is the so-called simple regret, because
it captures the quality of the eventually selected action.

\begin{definition} \textbf{Simple regret} of a policy $\pi$ at state $s$ is
defined as 
\begin{equation}
\SR(s, \pi) = \E_{a \sim \pi(s, \cdot)}\left[\max_{a' \in \mathcal{A}_s} Q(s, a') - Q(s, a) \right] \: ,
\label{eq:sr}
\end{equation}
where $\pi(s,\cdot)$ is a curried function whose domain is $\mathcal{A}$. 
\end{definition}

However, in the current setting, the $Q$-function is given by the environment
and is unknown. Therefore, at best, we can minimize our subjective estimate of
simple regret, which captures how much regret the agent expects to face if he
is to take an action right away, which we refer to as Bayesian simple regret.
Formally:
\begin{definition} We define the \textbf{Bayesian simple regret ($\BSR$)} of a policy $\pi$ at state $s$ given a sequence of past computations $\ot$ as
\begin{equation} 
\BSR(s; \pi, \ot) = \mathbb{E}_{a \sim \pi(s, \cdot)} \left[ \mathbb{E}[ \max_{a' \in \mathcal{A}_s} Z(s,a' | \ot) ] -\mathbb{E}[Z(s,a | \ot)] \right]\: ,
\end{equation}
where $\pi$ potentially takes $\ot$ into account.
\end{definition}
It should be clear that the policy $\pi$ that minimizes $\BSR$ is a
deterministic policy that assigns probability $1$ to the action $\arg \max_{a
\in \mathcal{A}_s} \mathbb{E}[Z(s,a | \ot)]$. Thus, we introduce the optimal Bayesian simple regret ($\BSRo$) as
\begin{align} 
\BSRo(s ; \ot) &= \min_\pi \BSR(s, \pi ; \ot) \\
&= \mathbb{E}[ \max_{a \in \mathcal{A}_s} Z(s,a | \ot)] - \max_{a \in \mathcal{A}_s} \mathbb{E}[Z(s,a | \ot)] \: . \label{eq:bsr} 
\end{align} 

For a given policy $\pi$, we would like to see the relation between $\BSR$ and
\SR. Recall that $Z(s,a | \ot) = Q(s,a) | \ot$. Thus, $\BSR$ is in a
sense the \emph{subjective} expected simple regret:
\begin{equation} 
\BSR(s, \pi ; \ot) = \E_{Q \sim Z | \ot}\left[ \underbrace{\E_{a \sim \pi(s, \cdot)} \left[ \
\max_{a' \in \mathcal{A}} Q(s,a') - Q(s,a)\right]}_{= \textrm{simple
regret}}  \right]\: ,
\end{equation}
where $Z|\ot = Z(\cdot, \cdot | \ot)$ is a random function.

\edited{

We can see that the agent can minimize the $\BSR$ at state $s$ by taking better
actions (i.e., controlling $\pi$) or performing better computations (i.e.,
controlling $\ot$). In reinforcement learning, where model of the environment
is not known, it might often be advantageous to take actions that seem to be
suboptimal---as the exploration induced by such actions might be beneficial in
the long-run. In planning problems, however, exploration of the state space is
not necessary nor beneficial as the model is already known. Therefore, the only
reasonable policy of selecting actions is to $\arg \max$ the expected values
obtained via the computations. Consequently, the only way to (non-trivially)
reduce $\BSR$ is to select better computations.

}

\begin{definition} A meta-policy is said to be \textbf{myopically optimal} if
and only if it performs a computation $\omega^*$ that minimizes the expected
$\BSRo$ given the current state $s$ and a sequence of computation outcomes $\ot$. That is, 

\begin{equation} \omega^* \coloneqq \arg \min_{\omega \in \Omega} \mathbb{E}_{o
\sim \omega}[ \BSRo(s; \ot \oplus o)] \: . 
\label{opt} 
\end{equation}
\end{definition}

\begin{proposition} $\VOC$-greedy is myopically optimal.
\label{prop:mab_myopt}
\end{proposition}

\begin{proof}
The optimal computation as defined in Equation~\ref{opt} remains unchanged if we subtract the current $\BSRo$. I.e.,
\begin{equation}
\omega^* = \arg \min_{\omega \in \Omega} \mathbb{E}_{o \sim \omega}[ \BSRo(s; \ot \oplus o) - \BSRo(s; \ot)] \: .
\end{equation}

The first terms of the $\BSRo$'s are equal,
\begin{equation}
\mathbb{E}[ \max_{a \in \mathcal{A}_s} Z(s,a | \ot)] \overset{!}{=}
\mathbb{E}_{o \sim \omega}\mathbb{E}[ \max_{a \in \mathcal{A}_s} Z(s,a | \ot
\oplus o)]
\end{equation}
, due to the law of total expectation. Hence we get,
\begin{eqnarray}
\omega^* &=& \arg \min_{\omega \in \Omega} \mathbb{E}_{o \sim \omega}[\max_{a} \mathbb{E}[Z(s,a | \ot)] -  \max_{a} \mathbb{E}[Z(s,a | \ot \oplus o) ]   ] \\ 
&=& \arg \max_{\omega \in \Omega} \underbrace{\mathbb{E}_{o \sim \omega}[ \max_{a} \mathbb{E}[Z(s,a | \ot \oplus o)]  - \max_{a} \mathbb{E}[Z(s,a | \ot)]]}_{= \VOC(\omega; s, \ot)}
\end{eqnarray}
Thus, $\VOC$ is the expected change in $\BSRo$; and a meta-policy that minimizes $\BSRo$ maximizes the value of computation.
\end{proof}

\edited{
\begin{proposition} $\VOCp$-greedy is myopically optimal.
\end{proposition}

\begin{proof} This is a direct corollary of Propositions~\ref{eq_voc} and \ref{prop:mab_myopt}.
\end{proof}
}
What about the asymptotic behavior as the agent performs infinitely many computations? 

\begin{proposition}If the sampling distributions are known, $\VOC$-greedy is asymptotically optimal in the sense that simple regret approaches $0$ in the limit of computations performed.
\label{asymp}
\end{proposition}
This is proved by Frazier et. al. \cite{frazier2009b} in the setting of knowledge gradient policies, results of which directly apply here. \edited{Here, we give an intuitive but incomplete explanation why this is the case.}

\begin{hproof}
There are two conditions in which $\VOC$-greedy might fail to be
asymptotically optimal: (I) the meta-policy halts early, before ensuring simple
regret is $0$; (II) the meta-policy does not halt, yet simple regret does not
approach $0$.

Note that $\BSRo$ equals zero if and only if there exists a policy whose simple
regret is zero. This is because SR is non-negative, and $\BSRo$ is the
subjective expectation of SR under the best policy.

Also recall that $\VOC$ of $\omega$ is the expected reduction in $\BSRo$
resulting from $\omega$. Thus, $\exists \omega: \VOC(\omega; s, \ot) > 0
\implies \BSRo(s; \ot) > 0$. \edited{The converse is also true for sampling
distributions with infinite support.}
Thus, the meta-policy does not halt
until $\BSRo$ hence the simple regret is $0$. The condition (II) can only
happen if an action is sampled finitely many times, which cannot happen for
probability distributions with infinite support, as there will always be a chance of a
sub-optimal action turning out to be the optimal one after a computation.
\end{hproof}

\edited{In addition to myopic and asymptotic optimality, finite error bounds can also be 
computed as done in \cite{frazier2009b}.}

Given our assumption about all the uncertainty being computational; 
posterior values become delta functions given enough computation. If a similar
approach is taken in a model-based scenario, where the agent has to learn the
statistics of the environment to make better decisions, the $Z$-function would
capture the value marginalized over empirical but not computational
uncertainties. \edited{Alternatively, the $Z$-function could account for both kinds of
uncertainties; however, this would bring added complexity---as regret cannot be
made zero solely via computations---and require a different formulation then 
we propose here.}

We should point out that asymptotic optimality is easy to obtain. Heuristic
policies such as uniform random sampling, or $\epsilon$-greedy also are
asymptotically optimal as they sample all the actions infinitely many times.
However, $\VOC$-greedy is the only policy that is both myopically and
asymptotically optimal\footnote{Besides the $n=1$ case, where $\VOC$-greedy and
$\VOCp$-greedy are equivalent}. What can we say about optimality for the cases
in between these two extremes? \edited{We have no optimality guarantees. In fact, computing the
optimal meta-policy that maximizes the cumulative computation value of $B$ many
computations is shown to be NP-hard in $B$ \cite{madani2003}. That being said,
optimality in two extremes suggests that the performance is likely good in
between as well?but likely not optimal. Therefore, myopic
and asymptotic optimality is the best we can obtain efficiently in general.}

In this section, we were only concerned with computations and actions regarding
the state $s$ occupied by the agent. We assumed the sampling distribution of
$Q(s,\cdot)$ is a given---which could be unrealistic in multi-state settings
(MDPs) as computing the sampling distribution might require actually solving
the problem. We address this problem in Section~\ref{sec:mdp}. Before that, we
direct our attention to an easier problem, namely the multi-armed bandit
problem.

\section{Computations in multi-armed bandits} 

In this section, we consider a more concrete setting, and motivate the
usefulness of $\VOC$ in multi-armed bandits, or equivalently, in discrete
Bayesian optimization. \edited{Here, we will focus on selecting the optimal
action to perform, rather than the optimal computation. We show that the
formalism behind the two are analogous and our $\VOC$ results are directly
applicable here. Consequently, we do not introduce any new ideas here and
readers who are not interested in a concrete example are encouraged to skip
this section. We should also mention that this problem has been studied in the
setting of the \emph{knowledge gradients} and similar results can be found in
\cite{frazier2009b}.} Below, we (re)introduce the multi-armed bandit problem we have discussed in Introduction before.

\begin{definition} 
\edited{In a \textbf{multi-armed bandit (MAB)} problem, there is a finite set of
actions $\mathcal{A}$, and a value function $Q: \mathcal{A} \rightarrow
\mathbb{R}$, which is unknown to the agent. At each time step $t$, the agent
takes an action $a_t \in \A$ and collects $r_t = Q(a_t) + \epsilon$ where $\epsilon$ is
sampled i.i.d. from a distribution with zero mean and potentially depends on
$a_t$. The goal is to minimize a measure concerning the expected simple regret---which is defined as $R(t) = \E_{a_t}[\max_{a \in \A} Q(a) - Q(a_t)]$. One can either aim to minimize this directly for some $t$, or its time sum, that is $\sum_{i=1}^n R(i)$ for some $n$.
}
\end{definition}

\begin{example} 
Consider the setting of a MAB problem, where $\epsilon \sim
\mathcal{N}(0, 1/\tau)$ i.i.d. for a known $\tau$ and $Q(a) \sim
\mathcal{N}(\mu_0, 1/\tau_0)$ i.i.d. for known $\mu_0$ and $\tau_0$. Let $\rt =
r_1, r_2, \dots, r_t$ be a sequence of returns yielded by sampling \emph{a specific action} $a$ and
$\hat{r}_t = \frac{1}{t} \sum_{i=1}^t r_i$. \edited{ Thus, at time step $t$, if the agent takes action $a$ and receives reward $r_t \in \mathbb{R}$, then we define the observation resulting from this action as $o_t = \langle r_t, a \rangle$.}
Then, the posterior is obtained by
\begin{equation}
Z(a|\ot) \sim \mathcal{N}\left(\frac{t \tau \hat{r}_t + \tau_0 \mu_0}{t \tau + \tau_0}, \:\frac{1}{t \tau + \tau_0} \right)  \: .
\label{zart}
\end{equation}
where $\mathcal{N}$ is the normal distribution. Then, the posterior predictive
distribution of a return sampled form $a$ is $Z(a|\ot) + \epsilon$.
We can see that the formulation for picking actions based on observations is
equivalent picking computations as we have done previously.
Let us consider whether equality asserted by the law of
total expectation holds as in Equation~\ref{tower2},
\begin{equation}
\E_{o \sim \omega_a}\left[\E[Z(a|\ot \oplus o)] \right] \overset{?}{=} \E[Z(a|\ot)] \: ,
\label{tower?}
\end{equation}
\edited{where $o$ is what the agent supposes to observe after performing action $a$. We use the notation $o \sim \omega_a$ instead of $o \sim a$ in order to keep the notation consistent with our $\VOC$ formulation.}
We have
\begin{align}
\E_{o \sim \omega_a}\left[ Z(a| \ot \oplus o) \right]& \sim \mathcal{N}\left(\frac{(t + 1) \tau \hat{r}_{t+1} + \tau_0 \mu_0}{(t+1) \tau + \tau_0}, \:\frac{1}{(t+1) \tau + \tau_0} \right) \\
& = \mathcal{N}\left(\frac{\tau( n\hat{r}_{t}  + r_{t+1}) + \tau_0 \mu_0}{(t+1) \tau + \tau_0}, \:\frac{1}{(t+1) \tau + \tau_0} \right) \: .
\end{align}
Given $r_{t+1}$ is sampled from the posterior predictive, $Z(a|\ot) + \mathcal{N}(0, 1/\tau)$, we can include its mean and variance,
\begin{align}
\E_{o \sim \omega_a}\left[Z(a|\ot \oplus o)\right] &\sim \mathcal{N}\left(\frac{\tau t\hat{r}_{t}  + \tau_0 \mu_0}{(t+1) \tau + \tau_0} + \frac{\tau}{(t+1) \tau + \tau_0} \frac{t \tau \hat{r}_t + \tau_0 \mu_0}{t \tau + \tau_0}, \right. \\
&\left. \frac{1}{(t+1) \tau +  \tau_0} +  \left( \frac{\tau}{(t+1) \tau + \tau_0}\right)^2 \left(\frac{1}{t \tau + \tau_0} + \frac{1}{\tau} \right) \right) \\
&= \mathcal{N}\left((\tau t\hat{r}_{t}  + \tau_0 \mu_0)\left(  \frac{1}{(t+1) \tau + \tau_0} + \frac{\tau/(t\tau + \tau_0)}{(t+1) \tau + \tau_0} \right) , \right. \\
&\left. \frac{1}{(t+1) \tau +  \tau_0} +  \left( \frac{\tau}{(t+1) \tau + \tau_0}\right)^2 \left(\frac{(t+1)\tau + \tau_0}{\tau(t\tau + \tau_0)}\right) \right) \\
&= \mathcal{N}\left((\tau t\hat{r}_{t}  + \tau_0 \mu_0)\left(  \frac{1}{(t+1) \tau + \tau_0} + \frac{\tau/(t\tau + \tau_0)}{(t+1) \tau + \tau_0} \right) , \right. \\
&\left. \frac{1}{(t+1) \tau +  \tau_0} +  \frac{\tau/(t\tau + \tau_0)}{(t+1) \tau + \tau_0} \right) \\
&= \mathcal{N}\left(\frac{t \tau \hat{r}_t + \tau_0 \mu_0}{t \tau + \tau_0}, \:\frac{1}{t \tau + \tau_0} \right) \: ,
\end{align}
which is the same distribution with that of $Z(a|\ot)$ as given in
Eq~\ref{zart}. Then,
\begin{equation}
Z(a|\ot) \overset{d}{=} \E_{o \sim \omega_a}\left[Z(a|\ot \oplus o)\right] \: ,
\end{equation}
where $\overset{d}{=}$ denotes a distributional equality, which in turn implies
\begin{equation}
\E[Z(a|\ot)]  = \E_{o \sim \omega_a}\left[\E[Z(a|\ot \oplus o)] \right] \: .
\label{zarteq}
\end{equation}
Thus the equality in Equation~\ref{tower?} holds and the probability estimates
are consistent. Furthermore, we can also see that
\begin{equation}
\VOC(\omega_a; \ot) \coloneqq \E_{o \sim \omega_a}\left[ \max_{a \in \mathcal{A}}\E[Z(a|\ot \oplus o)] \right] -  \max_{a \in \mathcal{A}}\E[Z(a|\ot)]  \ge 0\: ,
\label{eq:bandit_voc}
\end{equation}
due to Jensen's inequality. \edited{Note that, in this bandit settings, we do not perform computations but actions. We are evaluating the information obtained by performing $a$, which is usually referred to as the \emph{value of information}. However, because the underlying formulations are the same, we stick to $\VOC$.}

\edited{Let us consider the first term in Equation~\ref{eq:bandit_voc}. When considering the value of $t+1$th action, given $t$ many observations, we have
\begin{equation}
\E[Z(a|\ot \oplus o)] \sim \mathcal{N}\left(\E[Z(a|\ot)], \frac{1}{t \tau + \tau_0} - \frac{1}{(t+1) \tau + \tau_0}\right) \: .
\end{equation}
In a sense, computations add variability to the expectations of $Z$'s---which is good, as higher variability under convexity (i.e., $\max$ in our case) mean higher expectation. This might seem puzzling, as we would like to reduce the uncertainty by performing computations, which is actually the case as we have 
\begin{equation}
\Var \left[Z(a|o_{1:t+1})] \right] = \frac{1}{(t+1) \tau + \tau_0} \le  \Var \left[Z(a|\ot) \right] =  \frac{1}{t \tau + \tau_0} \: .
\end{equation}
In other words, computations reduce the total variance by shifting a fraction of the variance into the expectation, which subsequently is resolved by performing a computation.

Lastly, in this example, where $Q(a)$ are sampled independently
across $a$, then $\VOC$ can be calculated analytically, as it will be the expectation of a
truncated normal distribution. In the case of correlated $Q(a)$ samples with a known covariance, the analytical solution still exists \cite{frazier2009a, frazier2009b}; however, the involved computations are more costly.}
\label{ex_mab}
\end{example}

If the sampling distributions are not known, then $\BSR$ will not be equivalent
to expected simple regret. $\VOC$-greedy will still be myopically optimal with
respect to the \emph{assumed} distribution of $Q$ by construction. However, it
won't be guaranteed to minimize the expected simple regret as it is a function
of the true $Q$, which has an unknown distribution. In terms of asymptotic
optimality, the prior needs ensure that each arm will be sampled infinitely
many times, which is sufficient but not necessary.

\section{Computations in Markov decision processes} \label{sec:mdp}

We saw $\VOC$-greedy (and also $\VOCp$-greedy) are myopically and
asymptotically optimal in minimizing Bayesian simple regret in multi-armed
bandits. Eventually we would like to see if this
transfers to MDPs. Assume we treat the actions at the current state $s$ as
bandit arms. Can we apply the principles from the previous section to solve
this bandit problem?

The main obstacle is that in MDPs, action values are not solely given by the
environment, but are a product of both the environment and the agent's policy.
To complicate the matters further, the policy of the agent likely improves as
he performs more computations; thus, the action values should ideally reflect
the knowledge of his future self. To circumvent this problem, we introduce
static and dynamic values, which respectively capture how much the agent
should value an action if he can no longer perform any computations or if he
can perform infinitely many computation in the future.

\edited{
Before moving on, let us consider a puzzle which will yield insights into static and dynamic values.
As illustrated in Figure~\ref{optionality}, there are two rooms: the first room contains two boxes, and the second one contains five boxes. Each box contains an unknown amount of money. First you need choose a room, then you can select any of the boxes in the room---but peeking inside is not allowed---and collect the money. Which room should you choose? What if you could peek inside the boxes upon choosing the room? 

\begin{figure}[h]
\begin{center}
\includegraphics[width=0.8\textwidth]{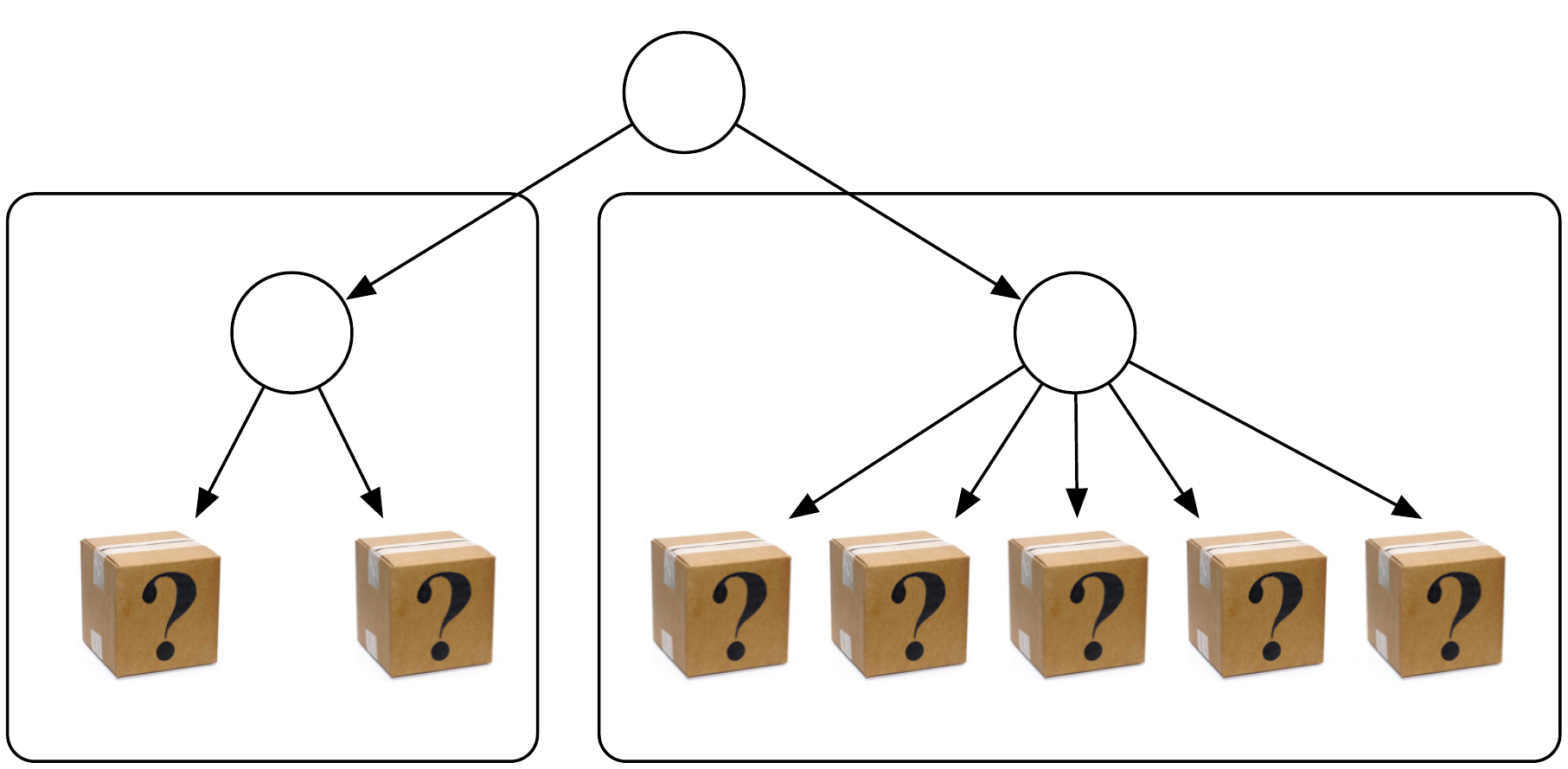}
\caption{Graphical illustration of the puzzle. There are two rooms, one containing two boxes, another one containing five boxes. There is an unknown amount of money in each box.}
\label{optionality}
\end{center}
\end{figure}

In the first case, it doesn't matter which room one chooses, as all the boxes are equally valuable in expectation, as no further information is given. Whereas, in the second case, choosing the second room, the one with five boxes, is the better option. This is because one can obtain further information by peeking inside the boxes---and more boxes mean more money in expectation, as one has the option to choose the best one. Formally, let $X = \{x_i\}_{i=1}^{n_x}$ and $Y = \{y_i\}_{i=1}^{n_y}$ be the sets of boxes of the first and the second room respectively. Assume all the boxes are sampled i.i.d. from an unknown distribution and $n_x < n_y$. Then, if one cannot peek inside the boxes, we have the equality of two rooms, as $\max_{x \in X} \E[x] = \max_{y \in Y} \E[y]$. Whereas, is one can peek inside all the boxes, then the $Y$-room is more valuable in expectation as $\E[\max_{x \in X} x] = \E[ \max_{y \in Y} y]$. Therefore, it is smarter to pick the room with more boxes, if peeking is allowed.

Alternatively, what if one can only reveal the contents of three boxes, which needs to be done before committing to a room? Or what if one gets to reveal one box before selecting a room, and reveal another one after selecting the room?

In this chapter, we address these questions in a principled way, through static and dynamic values. In the next chapters, we show that the answers we provide are directly applicable to MCTS---where one has imperfect information about action values, and gets to ``reveal'' more information by performing rollouts. 

}

 Using these, we redefine
computation values for MDPs. It turns out that $\VOC$-greedy is still
myopically and asymptotically optimal, whereas this is no longer the case for
$\VOCp$-greedy.

Before moving on to MDPs from MABs, we introduce an intermediate problem, where
the idea is to ``cut out'' a ``reduced-MDP'' by uniformly expanding in all
directions from the current state for a fixed number of steps\footnote{The
expansion does not have to be uniform, but we consider this scenario for
simplicity.} (i.e., actions). \edited{To this end, we introduce new value function measures that we use to relate the action values that lie on the ``cut''  to the current state of the agent.}

\begin{definition} Values of state-action pairs can be defined
recursively\footnote{As in the Bellman policy evaluation equation.}, which we refer to as the \textbf{$n$-step value}, and obtain as
\begin{equation}
Q^\pi_n(s, a) = \begin{cases}
Q^\pi_0(s,a) & \text{if $n = 0$}\\
\sum_{s'}\mathcal{P}_{ss'}^{a} \left[ \mathcal{R}_{ss'}^{a} +  \gamma \sum_{a' \in A_{s'}} \pi(s',a') Q^\pi_{n-1}(s', a') \right]&\text{else}\\
\end{cases} \: ,
\end{equation}
where $Q_0: \mathcal{S} \times \mathcal{A} \rightarrow \mathbb{R}$ is a measure
of how valuable frontier state-actions are.
\end{definition}

We will consider different assumptions concerning $Q_0$. They can belong to the
environment as in the expected rewards of bandit arms. Alternatively, the
spatial horizon of $n$ can be computational. Meaning, even though there are reachable
steps beyond $n$, the agent ignores them and treats the expected rewards of the
frontier actions as bandit arms. \edited{In either case, we assume that they are independent of agent's policy. Justification of this assumption is two-fold. First, the impact of the policy on action values diminishes as we go farther away from the current state. Second, as we discuss later, we could implicitly account for the policy at remote states by leveraging an adaptive sampling policy such as the UCT-policy for Monte Carlo tree search.}

\edited{
In practice, if we have a spatial horizon of $n$, we will compute $Q^\pi_n$ for the actions that are available at the current state, and $Q^\pi_{n-1}$ for the actions that are $1$-step reachable, and so on, until we get to the state-actions that lie on the cut, for which we will have $Q_0$.
}

The $n$-step value is convenient because it provides a trade-off between a MAB
problem and a full-fledged MDP. Let $n=0$, and $Q_0(s,a)$ be drawn from a known
sampling distribution. The agent does not know $Q_0(s,a)$, but can draw noisy
samples from it. \edited{Then the problem of finding the best computation to maximize the root action value here is equivalent to the problem of finding the best action to perform in a MAB problem, where the goal is to minimize the expected simple regret of the next time step.}
For $n>0$, the $n$-step value generalizes this
MAB problem to multiple states. Also for $n \rightarrow \infty$, we essentially
get the full Bellman backups that solves an MDP.

Note that, even if the estimates of $Q_0$ are highly unreliable, the estimates
of proximal state-actions (e.g., of high $n$) will be more reliable given the
$\gamma$ that shrinks the errors by $\gamma^n$ and $\sum$ operations which
``average out'' the noise of the frontier estimates. We now reintroduce
$n$-step optimal value function, leaving $\pi$ out of the equation:

\begin{definition} We define the \textbf{$n$-step optimal value} function as
\begin{equation}
Q^*_n(s, a) = \begin{cases}
Q_0(s,a) & \text{if $n = 0$}\\
\sum_{s'}\mathcal{P}_{ss'}^{a} \left[ \mathcal{R}_{ss'}^{a} +  \gamma \max_{a' \in A_{s'}} Q^*_{n-1}(s', a') \right]&\text{else}\\
\end{cases} \: ,
\label{nso}
\end{equation}
where $Q_0: \mathcal{S} \times  \mathcal{A} \rightarrow \mathbb{R}$ is a measure of how valuable frontier state-actions are.
\end{definition}

The motivation for this comes from the fact that, the structure of a Monte
Carlo Search Tree resembles that of an $n$-step optimal value equation. The
leaves of the tree offer imperfect estimates of action values, which are
propagated to the root of the tree in order to rank and select the best action
available at the root. One difference is that MCTS algorithms build trees
iteratively, whereas we assume full-expansion up to depth $n$. In fact, as a
result of this expansion, we end up with a weighted directed acyclic graph
(DAG) where the current state is the sink, and all $n$-step reachable nodes are
sources. Thus, our data structure is not a tree as in MCTS, but it is a graph,
which we refer to as a \textbf{search graph}. The crucial difference is that,
in a tree, each node can have a single parent; whereas in a DAG, a node can
have multiple parents. Therefore, a DAG is more (sample-wise) efficient than a
tree for evaluating action values, as the latter does not share the information
across branches \cite{childs2008}. Consequently, instead of a root node as in
MCTS, we have a \textbf{sink}, which is the state of the agent; and instead of
leaves, we have \textbf{sources}. Values, or uncertainties, flow from sources
to the sink in a sense. If a state-action pair is far from the sink, we refer
to it as being \textbf{distal}; and if it is close, as being \textbf{proximal}.

From now on, we assume we are given a search graph, and we would like to
perform computations, which are the noisy estimates of source values, in order
to minimize the $\BSRo$ at the sink. As before, our action value estimates are
probabilistic and are denoted by $Z$, such that $Z(s,a | \ot) = Q_0(s,a) |
\ot$.

\begin{definition} Given that we are now uncertain about the action values, we
define the \textbf{dynamic value} function as $\psi_n(s,a | \ot) \coloneqq
\mathbb{E}_{Z | \ot}\left[ \Upsilon_n(s,a | \ot) \right]$ where,
\begin{equation}
\Upsilon_n (s, a | \ot) \coloneqq \begin{cases}
Z(s,a | \ot) & \text{if $n = 0$}\\
\sum_{s'}\mathcal{P}_{ss'}^{a} \left[ \mathcal{R}_{ss'}^{a} +  \gamma \max_{a' \in A_{s'}} \Upsilon_{n-1}(s', a' | \ot) \right]&\text{else}\\
\end{cases} \: .
\label{psi}
\end{equation}
\edited{$\psi_n(s,a)$ captures how valuable $\langle s,a \rangle$ is, assuming the agent will have obtained all the information needed to completely eliminate his uncertainties about the frontier action values. The `dynamic' reflects the fact that the agent will likely change his mind about the best actions available at each states in the future; yet, this is reflected and accounted for in $\psi_n$.
}
\end{definition} 

This value is useful, because imagine you are forced to take an action right
away, after which you will have the luxury of deliberation. In this case, the
value of your action should take into account that the future version of you
will be wiser, which $\psi_n$ achieves. The major downside is that it cannot 
be computed efficiently in general. However, in the algorithms section,
we will introduce an efficient way of computing $\psi_n$ approximately.
Alternatively, we can utilize another measure, which captures the other
extreme, where the agent cannot obtain any new information.

%

\begin{definition} We define the \textbf{static value} function as
\begin{equation}
\phi_n (s, a | \ot) \coloneqq \begin{cases}
\mathbb{E}\left[ Z(s,a | \ot) \right]& \text{if $n = 0$}\\
\sum_{s'}\mathcal{P}_{ss'}^{a} \left[ \mathcal{R}_{ss'}^{a} +  \gamma \max_{a' \in A_{s'}} \phi_{n-1}(s', a' | \ot) \right]&\text{else}\\
\end{cases} \: .
\end{equation}
$\phi_n(s,a)$ captures how valuable $\langle s,a \rangle$ would be if the agent were to take $n$ actions before running any new computations. We also refer to $n$ as the \textbf{spatial horizon}.
\end{definition}
\edited{
In Figure~\ref{phi_psi}, we graphically contrast dynamic and static values, where the difference is the stage at which the expectation is taken. For the former, it is done at the level of the root actions; for the latter, at the level of the leaves.

\begin{figure}[h]
\begin{center}
\includegraphics[width=\textwidth]{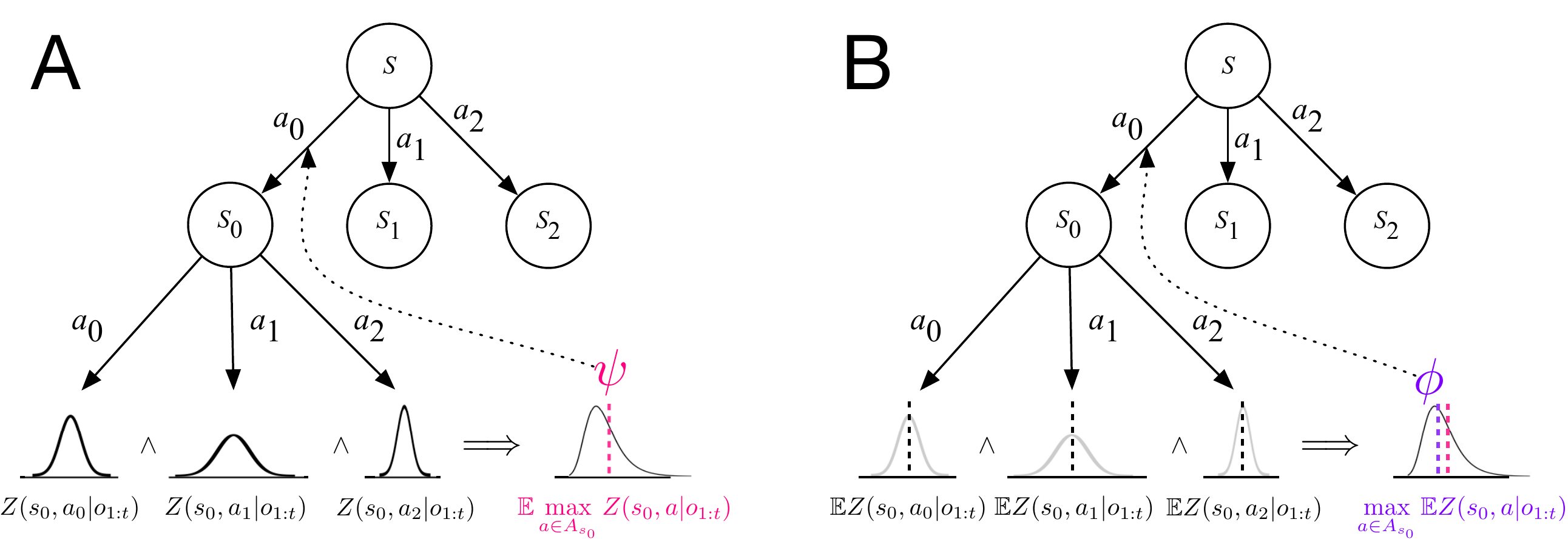}
\caption{Graphical illustration dynamic and static values. We ignore immediate rewards and the discounting for simplicity, and consider $2$-step values. In Panel~A, dynamic values (given by $\psi$) are obtained by calculating the \emph{expected maximum} of all state-action values (given by $Z$) lying $2$-steps away. Whereas, the static values (given by $\phi$) are obtained by calculating the \emph{maximum of expectations} of state-action values, as shown in Panel~B.}
\label{phi_psi}
\end{center}
\end{figure}
}

Note that $\psi_n(s,a | \ot)$ is a convex function of $Z(\cdot,\cdot |\ot)$ evaluated at the frontier nodes. Thus, let us denote $\psi_n(s,a | \ot)$ as $f(Z)$ where $f$ is a convex functional. Then we have $\phi_n(s,a | \ot) = f(\E Z)$. Consequently, by Jensen's inequality, we get 
\begin{equation}
\psi_n(s,a | \ot) \ge  \phi_n(s,a | \ot) \: .
\end{equation}

This shows that, regardless of the computation outcomes, i.e. $\ot$, the
dynamic value is at least as big as the static value, which is intuitive.

In addition, $\psi_n$ accounts for what we refer to as \textbf{optionality},
namely the fact that given the agent might gather more information in the
future, he should value the \emph{option} to change his mind (and his policy).
\edited{Flexible flight tickets---where one can
change the dates without paying a penalty---are more expensive because of
the optionality. By buying a flexible ticket to Sicily, you are paying for the option to
postpone your trip for a few days if your holiday turns out to be in risk of being ruined by
bad weather.}

So far we have only examined properties of $\psi_n$ and $\phi_n$
statically---given past computation outcomes $\ot$. Now, the effects of we
consider future computations, for which we need to consider two cases: (I) the
sampling distributions of $Q_0(s,a)$ and $\epsilon$ are known and stationary;
and (II) unknown and non-stationary.

\subsubsection{Known sampling distributions} 

Let us consider the case where the sampling distribution of $Q_0$, as well as
that of the noise, is known, but $Q_0$ itself is unknown. We can think of this
case as an MDP, where all state-action pairs $n$-step away from the current
state are stationary bandits, and we would like to sample the bandit arms in
order to determine the values of actions, which then would be propagated back
to the state occupied by the agent. In other words, all the reducible
uncertainty\footnote{In contrast to the irreducible uncertainty, which is given
by the environment, for instance, in terms of the stochasticity of state
transitions.} concerns the values of the frontier actions. Therefore, the agent
attempts gain certainty about those actions in an effort to improve his
decision quality at his current state.

In this setting, $Z(s,a)$ is the known sampling distribution of $Q_0(s,a)$,
hence the prior; and $Z(s,a | \ot)$ is the posterior distribution conditioned
by a sequence of computation outcomes $\ot$. Then we have,
\begin{equation}
\E_{Q_0 \sim Z}[Q^*_n(s,a)] =  \psi_n(s,a |  \langle \rangle) \: ,
\end{equation}

where $\langle \rangle$ is a null computation sequence, i.e., with no elements.
Furthermore, dynamic values are stable in expectation under any sequence of
future computations $\wk$, which would add $k$ more computation outcomes to the
outcomes existing in $\ot$. At this point, a computation consists of drawing a
sample as well as computing the updated $\psi_n$ and $\phi_n$ given the sample.

We can see that $\psi_n$ is consistent with respect to the new computations---meaning it is constant in expectation,
\begin{equation}
\psi_n(s,a | \ot) = \E_{\ok \sim \wk}[\psi_n(s,a | \ot \oplus \ok)] \: ,
\label{constancy}
\end{equation}
due to the law of total expectation. Note that the $k$ new computations can be applied in bulk without updating $\psi_n$ after every computation. Or $\psi_n$ can be updated after every computation, such that each computation would be based on the updated $Z$ resulting from the previous computations. Eq~\ref{constancy} holds for either approaches.
We can also relate it to the agent's expectation of $Q^*_n$,
\begin{equation}
\psi_n(s,a |  \ot) = \E_{Q_0 \sim Z | \ot}[Q^*_n(s,a)] \: .
\end{equation}
For $\phi_n$, we can assert that 
\begin{equation}
\E_{\ok \sim \wk}[\phi_n(s,a | \ot \oplus \ok)] \ge  \phi_n(s,a | \ot) \: ,
\label{phi_dom}
\end{equation}
which follows from Jensen's inequality. This suggests that the static values
cannot decrease in expectation after a computation sequence; yet, they are not
consistent as $\psi_n$ are. Combining these we get,
 \begin{equation}
\psi_n(s,a | \ot) \ge \E_{\ok \sim \wk}[\phi_n(s,a | \ot \oplus \ok)] \ge \phi_n(s,a | \ot) \: .
\end{equation}
In fact,
\begin{equation}
\psi_n(s,a | \ot) = \lim_{k \rightarrow \infty}\E_{\ok \sim \wk}[\phi_n(s,a | \ot \oplus \ok)]\: ,
\end{equation}
assuming a reasonable\footnote{More precisely, an asymptotically optimal one,
which is easy to realize.} meta-policy.

In other words, the value an agent hopes to reap from an environment by taking
an action---given he will have performed more computations in the future before
taking the action---falls between the static and dynamic values.
Therefore, we think these values form reasonable bases for computing
computation values, as they are independent of the policy as well as the future
policies.

\subsubsection{Bayesian simple regret and the value of computation in MDPs} 

We define Bayesian simple regret for MDPs, as we did for MABs, using the new
measures we introduced. Recall $\Upsilon_n(s,a | \ot)$ we introduced in
Equation~\ref{psi}, which is a random variable representing the uncertainty of
state-action values. Note that $\psi_n(s,a | \ot) \coloneqq \mathbb{E}_{Z |
\ot}\left[ \Upsilon_n(s,a | \ot) \right]$, thus $\mathbb{E}_{Z | \ot}\left[
\max_{a \in {\mathcal{A}_s}} \Upsilon_n(s,a | \ot) \right] \ge \max_{a\in
\mathcal{A}_s}\psi_n(s,a | \ot)$ by Jensen's inequality. The left-hand-side of
the inequality captures the value of $s$, if the agent could obtain perfect
information; whereas the right-hand-side captures the value of the best action
if the agent could completely resolve his uncertainty \emph{after taking the action}.
As such, we use $\Upsilon$ to formalize Bayesian simple regret. For a set of
past computation outcomes $\ot$, the $\BSRo$ at state $s$ is

\begin{equation} 
\BSRo(s; f, \ot)=  \E\left[\max_{a \in \mathcal{A}_s} \Upsilon_n(s,a | \ot)\right]- \max_{a \in \mathcal{A}_s} f(s,a | \ot) \: ,
\end{equation}
for $f$ is a real-valued function that values state-action pairs given
past information, in our case $f \in \{\phi_n,\psi_n\}$. Note that this definition is slightly different that the $\BSRo$ we have defined in Equation~\ref{eq:bsr} as we are now basing it on the static and dynamic values.

Similarly, we can adapt the definitions  of $\VOC$ and $\VOCp$ we have given before to MDPs simply by replacing $Z(s,a | \ot)$ with $\phi_n(s,a| \ot)$ or $\psi_n(s,a |  \ot)$,

\begin{align}
\VOC(\omega ; f, \ot, s) &= \mathbb{E}_{o \sim \omega}\left[\max_{a \in \mathcal{A}_s} f(s,a |\ot \oplus o) - \max_{a \in \mathcal{A}_s} f(s,a | \ot) \right] \label{vocmdp} \\
\VOCp(\omega ; f, \ot, s)& = \mathbb{E}_{o \sim \omega}\left[\max_{a \in \mathcal{A}_s} f(s,a | \ot \oplus o) - f(s, \arg \max_{a \in \mathcal{A}_s}  f(s,a | \ot) | \ot \oplus o) \right]\: \label{voc2mdp},
\end{align}
where $f \in \{\phi_n,\psi_n\}$ as well.

We introduce the algorithms $\VOC(\phi_n)$-greedy, $\VOC(\psi_n)$-greedy (same for $\VOCp$) which perform the computation that maximizes the specified $\VOC$ measure. If the computation budget is small, then selecting actions and computations w.r.t. $\phi_n$ should give better results, as it will be a better estimate of expected $Q^*_n$ conditioned on past and future computations. Whereas, if the budget is large, then we should value the uncertainties more, since the agent presumably will be able to gain additional value by resolving those uncertainties. Hence, $\psi_n$ should be a better measure.

\begin{proposition} $\VOC(\phi_n)$-greedy and $\VOC(\psi_n)$-greedy are still myopically optimal by construction in the sense that they maximize the expected decrease in $\BSRo(s; \phi_n, \ot)$ and $\BSRo(s; \psi_n, \ot)$ respectively.
\label{prop:myo_opt}
\end{proposition}
\begin{proof}
As we have shown before in a similar setting,
\begin{align}
\omega^* &\coloneqq \arg \min_\omega  \E_{o \sim \omega}[\BSRo(s; f, \ot \oplus o)]  \\
&= \arg \min_\omega  \E_{o \sim \omega}[\BSRo(s; f, \ot \oplus o)] - \BSRo(s; f, \ot) \\
&=  \arg \max_\omega \left[ - \E_{o \sim \omega} \left[\max_{a \in \mathcal{A}_s} f(s,a |\ot \oplus o) \right] + \max_{a \in \mathcal{A}_s} f(s,a | \ot) \right] \: ,
\end{align}
given $\E_{Z|\ot}\left[\max_{a \in \mathcal{A}_s} \Upsilon_n(s,a | \ot)\right] = \E_{o \sim \omega} \E_{Z|\ot}\left[\max_{a \in \mathcal{A}_s} \Upsilon_n(s,a | \ot \oplus o)\right]$. Thus,
\begin{equation}
\omega^* = \arg \max_\omega \VOC(\omega ; f, \ot, s) \label{voc_opt} \: .\\
\end{equation}
That is, selecting the computation that maximizes $\VOC$ minimizes the expected $\BSRo$, or more strongly, $\VOC$ of a computation is the expected decrease in the optimal Bayesian simple regret. 
\end{proof}
Then what can we say about $\VOCp$? Let us see if and when $\VOC$ and $\VOCp$ are equal. For this, the second terms in Equations \ref{vocmdp} and \ref{voc2mdp} need to be equal. That is,
\begin{align}
\mathbb{E}_{o \sim \omega}\left[f(s, \arg \max_{a \in \mathcal{A}_s}  f(s,a | \ot) | \ot \oplus o) \right] &\overset{!}{=} \max_{a \in \mathcal{A}_s} f(s,a | \ot)  \\
&= f(s,\arg \max_{a \in \mathcal{A}_s} f(s,a | \ot)  | \ot) \: .
\end{align}
This can only hold if the law of total expectation can be induced. If $f=\psi_n$, then we have 
\begin{align}
\mathbb{E}_{o \sim \omega}\left[\psi_n(s, \arg \max_{a \in \mathcal{A}_s}  \psi_n(s,a | \ot) | \ot \oplus o) \right] &= 
\mathbb{E}_{Z|\ot}\left[\Upsilon_n(s, \arg \max_{a \in \mathcal{A}_s}  \psi_n(s,a | \ot) | \ot \oplus o) \right]  \\
& = \mathbb{E}_{Z|\ot}\left[\Upsilon_n(s, \arg \max_{a \in \mathcal{A}_s}  \psi_n(s,a | \ot) | \ot) \right]  \\
&= \psi_n(s,\arg \max_{a \in \mathcal{A}_s} f(s,a | \ot)  | \ot) \: .
\end{align}

However, this does not hold for $f = \phi_n$ because the expectation of $\phi_n$ is not constant under computations as shown in Equation~\ref{phi_dom}. Thus we have,

\begin{align}
\VOC(\omega ; \psi_n, \ot, s) &= \VOCp (\omega; \psi_n, \ot, s)\\
\VOC(\omega ; \phi_n, \ot, s) &\ge \VOCp (\omega;  \phi_n, \ot, s) \label{diff_voc} \: .
\end{align}
Subsequently, we can make the following claims.
\begin{proposition} $\VOCp(\phi_n)$-greedy may not be myopically optimal.
\begin{hproof} We have shown that $\VOC(\phi)$-greedy is optimal. Computations selected by $\VOCp(\phi_n)$-greedy diverges from those selected by $\VOC(\phi)$-greedy given Equation~\ref{diff_voc} and as such is not optimal in general.
\end{hproof}
\end{proposition}

\begin{proposition} $\VOC(\phi_n)$-greedy is asymptotically optimal, but
$\VOCp(\phi_n)$-greedy may not be. \end{proposition}

\begin{proof} Optimality of $\VOC(\phi_n)$-greedy is due to the same conditions
that make $\VOC$-greedy optimal in MAB-like settings, as given by
Proposition~\ref{asymp}. We can show by counterexample that
$\VOCp(\phi_n)$-greedy may stop too early. Consider an agent in state $S$ of a
deterministic MDP, with no immediate rewards and $\gamma =1$. It forms a search
graph by uniformly expanding actions for $2$ steps. The black dots are the
frontier nodes, values of which the agent estimates by drawing samples.
Consider at time step $t$ of planning, the posterior distributions are as
described in Figure~\ref{counterr}.

\begin{figure}[htbp]
\begin{center}
\includegraphics[width=0.8\textwidth]{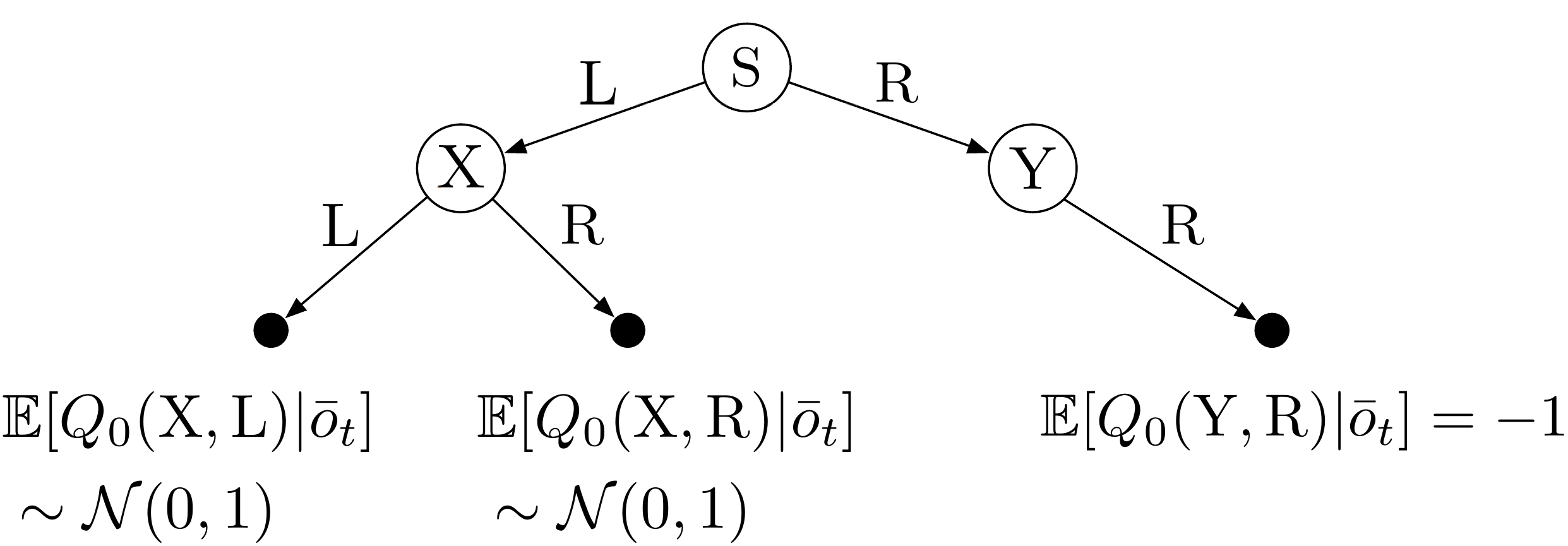}
\caption{A search graph, where $\VOCp(\phi_n)$-greedy stops early.}
\label{counterr}
\end{center}
\end{figure}

Assume $Z(\textsc{X}, \textsc{L} | \ot) \coloneqq \E[Q_0(\textsc{X},
\textsc{L}) | \ot]$ is orthogonal to $Z(\textsc{X}, \textsc{R} | \ot) \coloneqq
\E[Q_0(\textsc{X}, \textsc{R}) | \ot]$. Note that, $\phi_2(\textsc{S},
\textsc{L}) = \max_{a \in \{\textsc{L}, \textsc{R}\}} \E[Z(\textsc{X}, a)] = 0$
whereas $\phi_2(\textsc{S}, \textsc{R}) = -1$. Recall that $\VOCp > 0$ iff the
agent can change its mind about what the best action is. Current best action at
$\textsc{S}$ is $L$. However, sampling $\langle \textsc{X}, \textsc{L} \rangle$
or $\langle \textsc{X}, \textsc{R} \rangle$ \emph{once} cannot affect the best
action at $\textsc{S}$. That would require sampling both $\langle \textsc{X},
\textsc{L} \rangle$ and $\langle \textsc{X}, \textsc{R} \rangle$. Therefore,
the $\VOCp(\phi_n)$-greedy will stop early unlike $\VOC(\phi_n)$-greedy. This
can be interpreted as $\VOCp(\phi_n)$-greedy being \emph{lazy} as it defers
sampling until the point that it is useful. However, this laziness is costly
because it may result in choosing suboptimal actions. In this example, the
agent will not perform any computations and instead take the action
$\textsc{L}$ that will take him to $\textsc{X}$. However, there is a non-zero
probability that $\textsc{R}$ is the better action. Therefore, by taking action
$\textsc{L}$, the agent concedes an expected value loss. \end{proof}

\begin{proposition}$\VOC(\psi_n)$-greedy and $\VOCp(\psi_n)$-greedy are both myopically and asymptotically optimal.
\end{proposition}
\begin{hproof}
\edited{In Proposition~\ref{prop:myo_opt}, we established the myopic optimality of $\VOC(\psi_n)$-greedy. It is also asymptotically optimal due to the same conditions before---all frontier actions will be sampled infinitely many times. Both $\VOC(\psi_n)$-greedy and $\VOCp(\psi_n)$-greedy select the same computations. As such, both are myopically and asymptotically optimal.
}
\end{hproof}

\subsubsection{Unknown sampling distributions}

Let us briefly consider a more realistic case, where the sampling
distributions are unknown. Consider the MCTS algorithms proposed in
\cite{hay2012, tolpin2012}, where the agent \emph{treats} the root state as a
MAB problem such that pulling an arm is equivalent to \edited{sampling a roll-out
directly from that action}\footnote{\edited{This assumption is violated if the sampling is done via an adaptive process such as UCT, which is what the authors use to sample the ``arms''.}}.
If the rollout policy is fixed, then the
returns of the rollout will be sampled from a stationary but unknown
distribution. In this case, using the $\VOC$-greedy with weak priors
should work well in practice, for which we also provide evidence in the
experiments section.
 \edited{
 On the other hand, consider the case where the sampling procedure is adaptive. For
example, when sampling an arm is equivalent to using UCT---which
is an adaptive policy---as a subroutine. 
Then the expectation of the
sampling distribution shifts from the action values of the rollout
policy\footnote{Typically uniform random.}, to $Q^*$.} Note that, with each
rollout, we would expect the expectation of rollout returns to increase
\emph{in expectation}, even though they might decrease. In this case, if a weak
prior that assumes a stationary sampling distribution is used, $\phi_n$ can be
used to lower bound the expectation of $Q^*$ given the observations. However,
because of the upward drift in $Q$, upper bounding via $\psi_n$ has no
theoretical guarantees. Possible practical solutions include using a sliding
temporal window and taking the last $N$ observations into account and
exponentially discounting the past experiences, both of which are utilized in
the context of multi-armed bandit algorithms \cite{auer2002, kocsis2006b} for
estimating the empirical mean returns. However, using the same methods in a
Bayesian settings requires more care.

More generally, we know that as $n \rightarrow \infty$, $Q^*_n$ and thus
$\phi_n$ will approach $Q^*$ \edited{for any $n$---assuming sampling is done by an asymptotically optimal policy such as UCT.} 
We can also assert that as $t \rightarrow
\infty$, $\phi_n$ will approach $Q^*$. This is because all frontier nodes will
be sampled infinitely many times, and if they are sampled with UCT, then their
values will converge to $Q^*$, and thus, the proximal nodes will converge as
well. \edited{Furthermore, existence of finite sample error bounds for knowledge gradients \cite{frazier2009b} suggests that similar analysis can be done for $\VOC(\phi_n)$ and $\VOC(\psi_n)$-greedy policies, but this remains as a potential further work.}
In the experiments section, we will provide empirical evidence suggesting
that $\VOC$-greedy performs well even when the sampling distributions are
non-stationary and unknown.

\section{Discussion}
In this chapter, we introduced the value of computation ($\VOC$), and argued that maximizing it results in myopically and asymptotically optimal policies. More importantly, we provided a method to calculate $\VOC$'s in Markov decision processes that circumvents the problem of state-action values depending on the \emph{future} policy of the agent, which is unknown during the planning process. Our method operates on a MDP-like structure that is smaller than the original MDP. This yields desirable properties; however, is computationally costly and is not realistic from a neuroscientific or cognitive point of view.

\label{ch:voc}

\chapter{Computing the Value of Computation}

\begin{chapquote}{Unknown Author, \emph{Wisdom of Solomon}}
``The corruptible body presseth down the soul, and the earthly tabernacle weigheth down the mind that museth upon many things.''
\end{chapquote}

In the previous section, we proposed the $\VOC(\phi_n)$-greedy and
$\VOC(\psi_n)$-greedy algorithms by providing \emph{what} they compute
without addressing  \emph{how}. The latter is the topic of this section.

We first consider the case where state transitions are deterministic for
all state-action pairs. We then generalize the formulation to the general, stochastic
case. While describing these algorithms, we will appeal to an abstract
notion of computation as an operation that provides information about
action values. The detailed realization of the computations is left to
the next chapter, which uses the framework of Monte Carlo tree search
methods. 

\section{Deterministic state transitions}

\begin{figure}[htbp]
\begin{center}
\includegraphics[width=0.7\textwidth]{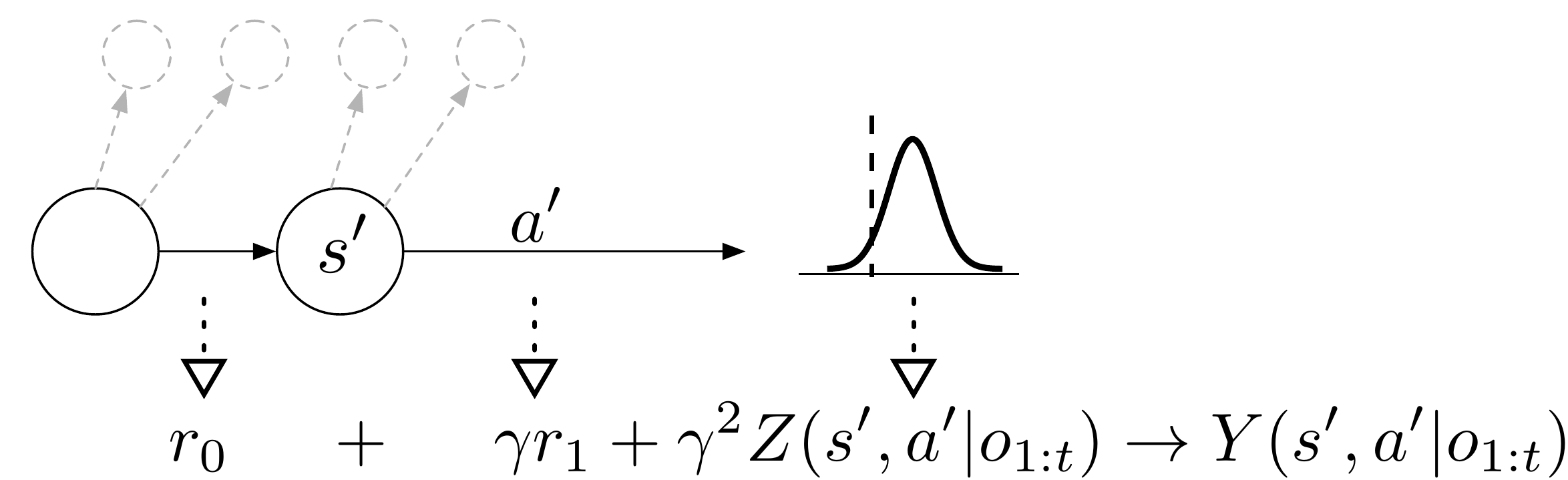}
\caption{$Y$ is composed of discounted combinations of immediate rewards, and the $Z$-function evaluated at the frontier node.}
\label{yz}
\end{center}
\end{figure}

If the state transitions are deterministic, then we can further simplify
the $n$-step optimal values defined in Eq~\ref{nso} by eliminating the
expectations over state transitions (i.e., $\mathcal{P}_{ss'}^a$). Then,
the recursively defined optimal values, together with static and
dynamic values, can be collapsed into a single $\max$, which is over
the source nodes, and a single $\E$ which is over the scaled and shifted
$Z$-values of the frontier nodes. The scaling is due to the discounting
by $\gamma^d$, where $d$ is the depth of the corresponding frontier
node; and shifting is due to the expected discounted immediate rewards
along the path from the
sink to the frontier node. Therefore, it is convenient to introduce a
new function, $Y$, which is the shifted and scaled $Z$. Then, we can
rewrite the optimal value equation for a root state action $\langle s,a \rangle$ as,

\begin{align}
\psi_n(s,a | \ot) &= \E\left[\max_\nsr Y(s', a' | \ot)\right] \label{psi_simple} \\
\phi_n(s,a | \ot) &= \max_\nsr \E\left[Y(s', a' | \ot)\right] \label{phi_simple} \: ,
\end{align}
where $Y(s',a' | \ot) \coloneqq r_0 + \gamma r_1 + \gamma^2 r_2 + \dots +
\gamma^d Z(s',a' | \ot)$, where $r_i$ denote the expected immediate reward
of depth $i$ that lies between the sink node and the given source (i.e.,
$\langle s,a \rangle$ here). We illustrate the relationship between $Z$
and $Y$ graphically in Figure~\ref{yz}.

It is easy to evaluate $\phi_n$, as it requires finding the maximum of expected $Y$ values. Whereas, $\psi_n$ requires finding the expected maximum of $Y$'s, as such, is harder. Even in the case of $Y$'s distributed i.i.d. according to a Normal distribution, we have no exact closed form solution. Therefore, $\psi_n$ needs to be computed approximately, for instance, by resorting to Monte Carlo sampling or variational methods. In the following subsection, we put forward an alternative method not only for approximating $\psi_n$, but also for approximating $\VOC(\psi_n)$ efficiently.

\subsection{Approximating $\psi_n$}

We utilize a bound by Lai and Robbins \cite{lrr76} enable us to get a
handle on $\psi_n$. This asserts,
\begin{align}
\psi_n(s,a | \ot) &= \mathbb{E} \left[\max_\nsr Y(s', a' | \ot) \right]  \\
&\underbrace{\le c + \sum_\nsr \int_c^\infty \left[1 - F^{s'a'}(x) \right] dx}_{\Scale[1.5]\lambda^{\Scale[1.1]{sa}}_{\Scale[1.1]{\ot}}} \label{eq:lrr}
\end{align}
where $F^{s'a'}$. This bound does not assume independence and holds for any correlation
structure by assuming the worst case. 

\edited{Equation~\ref{eq:lrr} is true for all $c$. However, the tightest bound is obtained
by differentiating the right side with respect to $c$, and setting it's derivative to zero, which in turn yields
$$\sum_\nsr \left[1 - F^{s'a'}(c) \right] = 1 \: .$$ Thus, the optimizing $c$ can be obtained via line search methods.
If $Y(s',a' | \ot)$ are distributed according to a Normal distribution,
then we can eliminate the integral},
\begin{equation}
\lambda^{sa}_{\ot} = c + \sum_\nsr \left[ (\sigma^{s'a'})^2 F^{s'a'}(c) + (\mu^{s'a'} - c) \left[1 - F^{s'a'}(c) \right] \right] 
\label{eq:lsa} \: .
\end{equation}

\subsection{Upper Expectation Bound (UEB) Algorithm}

We now introduce our first algorithm, which leverages the upper bound
given by $\lambda^{sa}_{\ot}$ for computing $\VOC(\psi_n)$. This can be naively done by leveraging the bounds we obtained for quantifying the value of a computation by 
\begin{equation}
\VOC(\omega; \psi_n, \ot, s) \approx \E_{o \sim \omega}[ \max_{a \in \As} \lambda^{sa}_{\ot \oplus o}]  - \max_{a \in \As} \lambda^{sa}_{\ot} \: .
\end{equation}
However, this is a very costly evaluation---\edited{even in the case of Normally distributed frontier node values}---
  because of the expectation. Instead, we can make the following
  approximation to reduce the burden by treating the time as a continuous variable momentarily, \begin{equation}
\lambda^{sa}_{\ot \oplus o} \approx \lambda^{sa}_{\ot} + \eta \frac{d \lsa}{d t} \: ,
\end{equation}
for some $\eta \in \mathbb{R}$. As we will see later, the exact value of $\eta$ will be insignificant.
Then,  we can now look at how a $\lsa$ (and thus $\max_a \lsa$) is expected to change locally with respect to $t$, by evaluating $d \lsa / dt$.

Assume $Y(s,a | \ot) \sim \mathcal{N} \left(\mu^{s'a'}_{\ot}, (\sigma^{s'a'}_{\ot})^2 \right)$; that is, $\mu^{sa}_{\ot}$ and $(\sigma^{sa}_{\ot})^2$ are the posterior mean and the variance respectively. Then, we have
\begin{equation}
\frac{d \lsa}{d t} = \sum_{\nsr} \left[ \frac{\partial \lsa }{\partial \sigma^{s'a'}_{\ot}} \frac{d \sigma^{s'a'}_{\ot}}{d t} +  \frac{\partial \lsa }{\partial \mu^{s'a'}_{\ot}} \frac{d \mu^{s'a'}_{\ot}}{d t} \right] \: .
\end{equation}

At this point, we could assume an independent conjugate Normal prior over $Q^*(s,a)$  as in Example~\ref{ex_mab}.  Then, a computation would only affect a single frontier node. That is, in this independent case all $d \mu^{sa}_{\ot}/dt$ and $d \sigma^{sa}_{\ot}/dt$ will be zero besides the one that is being sampled. Therefore, instead of taking derivatives with respect to $t$, we take with respect to $n^{s'a'}$ which is the number of times $\langle s',a' \rangle$ is sampled, thus $n^{s'a'} \le t$.
Then, as we have done in Example~\ref{ex_mab} posterior means and
variances can be obtained as,
\begin{align}
\mu^{s'a'}_{\ot} &= \frac{n^{s'a'} \tau \hat{r}^{s'a'}_t + \tau_0 \mu_0}{n^{s'a'} \tau + \tau_0} \: , \\ 
\sigma^{s'a'}_{\ot} &= 1 / \tau^{s'a'}_{\ot} = n^{s'a'} \tau + \tau_0 \: ,
\end{align}
where $\hat{r}^{s'a'}_t = \frac{1}{n^{s'a'}} \sum_{i=1}^{n_{s'a'}} r^{s'a'}_i$ is the mean of the sampled obtained from $\langle s', a' \rangle$, $\mu_0$ and $\tau_0$ are the prior mean and precision, and $\tau$ is the sample precision.
Subsequently, we obtain the following,
\begin{align}
\frac{\partial \lsa }{\partial \sigma^{s'a'}_{\ot}} &=  \frac{1}{\sqrt{2\pi}} \exp \left( \frac{-\left( \mu^{s'a'}_{\ot} - c\right)^2}{2\left(\sigma^{s'a'}_{\ot}\right)^{2}}  \right) \\
\frac{d \sigma^{s'a'}_{\ot}}{d n^{s'a'}} &= - \frac{\sigma \sigma_{0}^{3}}{2 \left(n^{s'a'} \sigma_{0}^{2} + \sigma^{2}\right)^{\frac{3}{2}}}\\
\frac{\partial \lsa }{\partial \mu^{s'a'}_{\ot}} &= \frac{1}{2} \left(1 +  \operatorname{erf}{\left (\frac{\sqrt{2} \left(\mu^{s'a'}_{\ot} -c \right)}{2 \sigma^{s'a'}_{\ot}} \right )} \right) \\
\frac{d \mu^{s'a'}_{\ot} }{d n^{s'a'}} &= \frac{\sigma^{2} \sigma_{0}^{2} \left(- \mu_{0} + \hat{r}^{s'a'}_t\right)}{(n^{s'a'})^{2} \sigma_{0}^{4} + 2 n^{s'a'} \sigma^{2} \sigma_{0}^{2} + \sigma^{4}}
,
\end{align}
combining these, for a computation $\omega^{s'a'}$, that provides
information about $\langle s', a' \rangle$, we get
\begin{equation}
\frac{\partial \lsa}{\partial n^{s'a'}} = \frac{\partial \lsa }{\partial \sigma^{s'a'}_{\ot}}
\frac{d \sigma^{s'a'}_{\ot} }{d n^{s'a'}}
+  \frac{\partial \lsa }{\partial \mu^{s'a'}_{\ot}} \frac{d \mu^{s'a'}_{\ot}}{d n^{s'a'}}\: .
\label{eq:del_lsa}
\end{equation}
\edited{Here, we consider the partial derivative of $\lsa$ with respect to $n^{s'a'}$, because there are potentially multiple such $\langle s', a' \rangle$ that can impact $\lsa$.}
Broadly, $\frac{\partial \lsa}{\partial n^{s'a'}}$ enables us to quantify the impact of sampling from a frontier node on root actions,  when $s$ is the root state, $a$ is a root action, and $\langle s', a' \rangle$ is a frontier state-action. Subsequently, we can utilize the following approximations for a computation $\omega^{s'a'}$ that provides information about $\langle s,a \rangle$,
\begin{align}
\VOC(\omega^{s'a'}; \psi_n, \ot, s) &\approx \E_{o \sim \omega^{s'a'}}[ \max_{a \in \As} \lambda^{sa}_{\ot \oplus o}]  - \max_{a \in \As} \lambda^{sa}_{\ot} \\
& \approx \E \left[ \max_{a \in \As} \left[ \lsa + \eta \partial \lsa / \partial n^{s'a'} \right] \right] -  \max_{a \in \As} \lsa \\
& =  \max_{a \in \As} \left[ \lsa + \eta \partial \lsa / \partial n^{s'a'} \right] -  \max_{a \in \As} \lsa \: , \label{vocapr}
\end{align}
setting $\alpha = \max_{a \in \As} \lsa$, we can further simplify by the following approximation
\begin{equation}
\max_{a \in \As} \left[ \lsa + \eta \partial \lsa / \partial n^{sa} \right] \approx \lsalpha + \eta \partial \lsalpha / \partial n^{s'a'} \: ,
\end{equation}
which we justify later.
Inserting this into Equation~\ref{vocapr}, we get
\begin{equation}
\VOC(\omega^{s'a'}; \psi_n, \ot, s) \approx \eta \partial \lsalpha / \partial n^{s'a'} \: ,
\end{equation}
where $\eta$ term can be ignored for $\arg \max$-ing the $\VOC$
approximation as it is a constant \edited{for the current state $s$}.
  This equation is intuitive, as it suggests, one could find the most promising root action---that is, $\alpha = \max_{a \in \As} \lsa$---and then find the $\langle s', a' \rangle$ that will have the biggest impact of $\lsalpha$ by $\arg \max$-ing $\partial \lsalpha / \partial n^{s'a'}$. However, $\lsalpha$ is an upper bound. As such, its maximization is not well justified. In order to remedy this, we utilize the \emph{optimistic initialization} heuristic that is widely used in reinforcement learning by initializing the $Z$-valued (or $Y$-values) optimistically such that $\partial \lsalpha / \partial n^{s'a'}$ is non-positive in expectation, i.e., $\E_{\ot}\left[ \partial \lsalpha / \partial n^{s'a'}  \right] \le 0$. 
In fact, it follows from Equation~\ref{eq:del_lsa} that $\partial \lsa/
\partial n^{s'a'} < 0$ if $\hat{r}^{s'a'}_t < \mu_0$ suggesting that we
would expect dynamic values to decrease with each computation if the
prior is sufficiently optimistic.

\begin{figure}[htbp]
\begin{center}
\includegraphics[width= \textwidth]{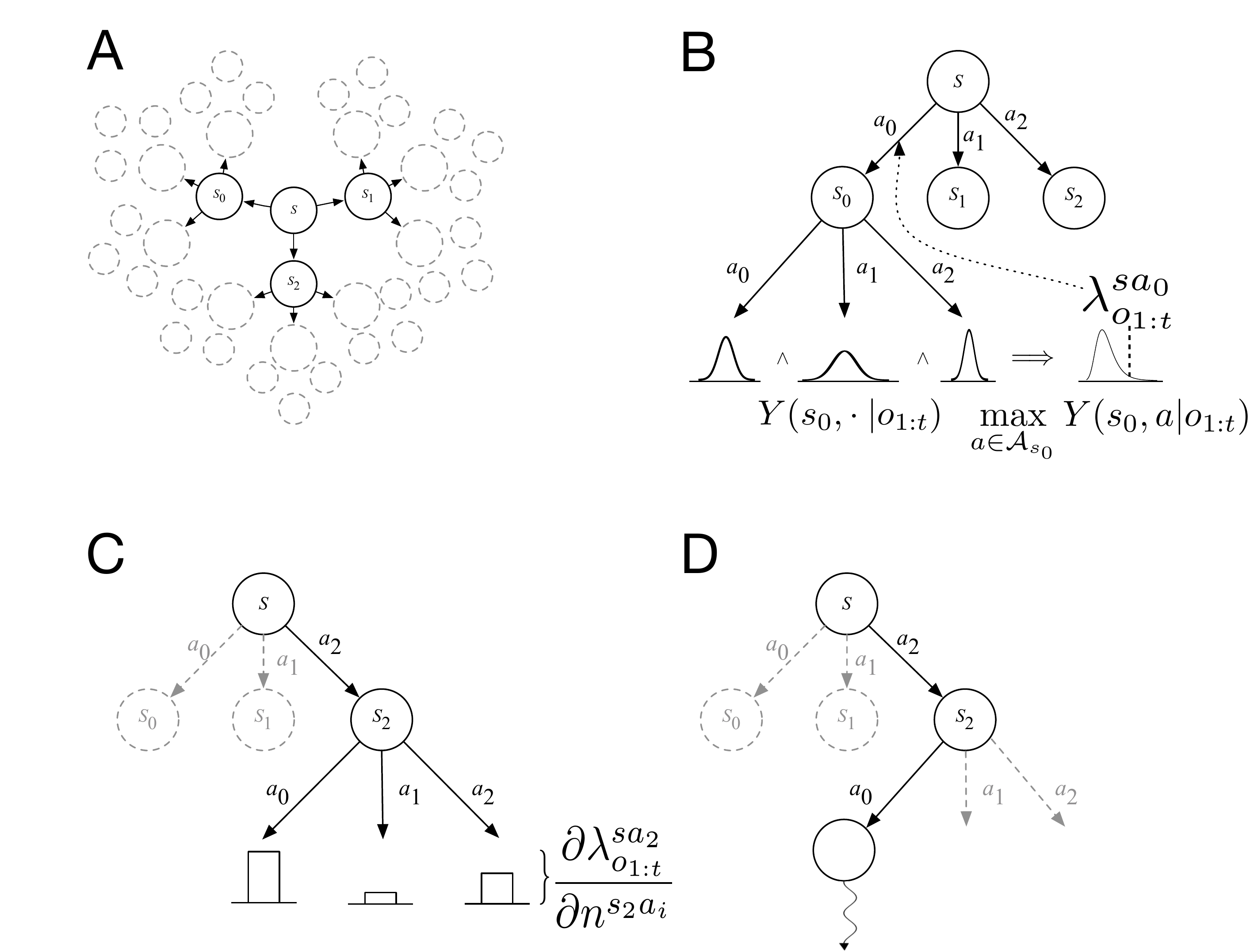}
\caption{Schematic description of the Upper Expectation Bound algorithm for an environment with deterministic state transitions. A: A partial search graph (or tree) is formed by expanding the current state $s$ along all directions for $n$ steps. In this illustration, $n=1$ for simplicity. B: $Y$-values on the frontier are propagated up to the root actions. Using these values, upper bounds for dynamic values of root actions are calculated, which are denoted with $\lambda$. C: The root action with the highest potential is determined, which is $a_2$, and then the frontier node that will have the highest impact on $\lambda^{s a_2}_{\ot}$ is selected, which is $\langle s_2, a_0 \rangle$. D: A computation is performed upon $\langle s_2, a_0 \rangle$ to obtain better information. In the case of Monte Carlo tree search methods, which we introduce later, this computation would involve performing a roll-out (i.e., trajectory simulation).}
\end{center}
\label{fig:ueb}
\end{figure}

Putting these together, we now put forward the Upper Expectation Bound (UEB) algorithm, which functions in four steps as illustrated in Figure~\ref{fig:ueb}

\subsection{$\VOC(\phi_n)$-greedy policy}
For the $\VOC(\phi_n)$-greedy policy, we need to compute $\max_{\nsr} \E[ Y(s',a' | \ot)]$ and $\max_{\nsr} \E[ Y(s',a' | \ot \oplus o)]$. The former is easy, as it is composed of picking the highest posterior mean.  The latter is also relatively to compute if $Y$-function is jointly Normally distributed, whether be independent as in Example~\ref{ex_mab} or be dependent as we discuss later on in this chapter.

\section{Stochastic state transitions}
\label{sec:stoc}

\edited{This analysis assumed deterministic transitions so far. If the transitions are stochastic, then we cannot collapse $\psi_n$ and $\phi_n$ into single $\E$ and $\max$ operators as we have done in Equations~\ref{psi_simple} and \ref{phi_simple}. This is because in the $n$-step optimal value equation, each step involves an expectation over the state transition probabilities, and a $\max$ over actions---unlike in the deterministic settings, where we only had nested $\max$'s, which could be collapsed into a single $\max$. In this stochastic case, $\phi_n$ can be computed by propagating the expected leaf values up the graph, which is equivalent to the backward induction method in game theory. For computing $\psi_n$, the same procedure can be applied by combining backward induction with sampling leaf values and averaging. 

Alternatively, we can approximate $\psi_n$ by using the upper bound given by,
\begin{equation}
\psi_n(s,a | \ot) \le \E_{\overline{T}} \left[ \E_Y \left[\max_{\nsr \textrm{ under }\overline{T}}  Y(s', a' | \ot)\right]  \right] \: .
\end{equation}
where $\overline{T}: \mathcal{S} \times \mathcal{A} \rightarrow \mathcal{S}$ is a deterministic transition function distributed according to $\mathcal{P}^a_{s \cdot}$, such that $\forall s \in \Sc, \forall a \in \A : \: \overline{T}(s,a) \sim \mathcal{P}^a_{s\cdot}$. Then, we can replace the inner expectation with the LR bound and obtain,
\begin{equation}
\psi_n(s,a | \ot) \le \E_{\overline{T}} \left[ \lsa(\overline{T})  \right] \: ,
\end{equation}
where 

\begin{equation}
\lambda^{sa}_{\ot}(\overline{T}) = c + \sum_{\nsr \textrm{ under }\overline{T}} \left[ (\sigma^{s'a'})^2 F^{s'a'}(c) + (\mu^{s'a'} - c) \left[1 - F^{s'a'}(c) \right] \right] \: .
\end{equation}

}

%
%
%
%

\section{Dealing with correlations}
Assuming independent sources simplify the $\VOC$ computations greatly. However, if correlations do exists among the optimal state-action values, treating them as independent will result in a unnecessarily high sample complexity. In other words, we can exploit correlations to learn action values with a fewer number of samples. 

This can be realized by, for instance, assuming a Gaussian process
prior, or a conjugate multivariate Normal prior over the joint distribution of frontier state-action values. Let $q$ be a vector composed of $Q^*$-values of the frontier nodes, whose elements are denoted with a superscript, e.g. $q^i$. Assume $q \sim \mathcal{N}(\mu_0, \Sigma_0)$ where $\mu_0$ is a prior mean vector and $\Sigma_0$ is a covariance matrix with elements $\mathbf{K}^{ij} = \mathrm{Cov}(q^i, q^j)$. As before, computations allow noisy observations, e.g., for time $t$, we can observe $r^i_{t} = q^i + \epsilon_t^i$ where $\epsilon_t^i \sim \mathcal{N}(0, \tau_i^{-1})$ i.i.d.. The posterior distribution $q | \ot$ can be obtained efficiently using the multivariate conjugate Normal prior. 

Specifically, we obtain the posterior $q | \ot \sim \mathcal{N}(\mu_t, \Sigma_t)$ where
\begin{align}
\mu_t &= \mu_{t-1} + \frac{r_t^i - \mu^i_t}{\tau^i + \Sigma_{t-1}^{ii}} \Sigma_{t-1}^{:i} \: ,\\
\Sigma_t &= \Sigma_{t-1} - \frac{\Sigma_{t-1}^{:i} (\Sigma_{t-1}^{:i})^T}{\tau^i + \Sigma_{t-1}^{ii}} \: ,
\end{align}
where $\Sigma_{t-1}^{:i}$ denotes the $i$th column of $\Sigma_{t-1}$.
If $\Sigma_0$ is constructed using a parametric kernel function, its
parameters can be optimized, \edited{for example by maximizing the likelihood function via gradient descent.}

Note that, the posterior obtained here is equivalent to the Gaussian
process (GP) posterior. If there are $N$ many frontier nodes, the joint
conjugate multivariate Normal posterior can be computed in
$\mathcal{O}(N^2)$, whereas the GP formulation requires $\mathcal{O}(t^3
+ N t^2)$ steps, as the fitting requires inverting a $t \times t$
matrix, and getting the posterior means and variances of $N$ nodes
requires an additional $N t^2$ steps. Therefore, there is no advantage
to using GPs unless $t \ll N$.

Alternatively, instead of optimizing the parameters of $\Sigma_0$, one can also set a conjugate prior for $\Sigma_0$ by assuming it is distributed according to the Wishart distribution.

Recall that the $Y$-function is the posterior of the $Q^*$-function given $\ot$, corrected by $\gamma$ and immediate rewards. We showed that we can update $Y$ efficiently, but we also need to be able to compute $\VOC$'s efficiently. For $\phi_n$, we need to evaluate 
\begin{equation}
\E_o \left[ \max_{\nsr} \E[Y(s',a' | \ot \oplus o)] \right] \: .
\label{eq:phi_eval}
\end{equation}
Here, the maximum is over a random vector, which is jointly Gaussian. However, all the randomness is caused by $o \in \mathbb{R}$. Thus, the joint posterior lies on a single dimensional manifold, 
where Eq~\ref{eq:phi_eval} is piece-wise linear in
$o$ and thus its expectation can be computed efficiently in
$\mathcal{O}(N^2 \log N)$ \cite{frazier2009b}. However, $\E_Y \left[
\max_{\nsr}Y(s',a' | \ot \oplus o) \right]$ is not piecewise-linear with
respect to $o$. Thus, we need to resort to approximate methods, such as
Monte Carlo sampling or variational inference for computing the
$\VOC(\psi_n)$-greedy policy. UEB, which approximates the
$\VOC(\psi_n)$-greedy policy, could also exploit the correlation in
principle. However, the derivation of gradient equations remains as a
potential direction for future work.

\chapter{Value of Computation for Planning}
\begin{chapquote}{Marcus Aurelius, \textit{Meditations}}
``You must compose your life action by action, and be satisfied if each action achieves its own end as can be.''
\end{chapquote}

In this chapter, we give a concrete realization of $\VOC$-greedy policies based on Monte Carlo tree search (MCTS). To this end, we first discuss MCTS in further detail first. Then, we derive new MCTS algorithms that value actions based on our theory of static and dynamic values, and select $\VOC$-maximizing simulations (i.e., rollouts). Finally, we compare our approach to other popular or relevant MCTS methods.

\section{Monte Carlo tree search (MCTS)}
\label{sec:mcts}

\edited{MCTS algorithms function by incrementally and stochastically building a
search tree to approximate state-action values. This incremental growth prioritizes 
the promising regions of the search space by directing the growth of the tree towards
high value states.
To elaborate, a tree policy is used to traverse the search tree and select a node which
is not fully expanded---meaning, it has immediate successors that aren't included in the tree. Then, the node
is expanded once by adding one of it's unexplored children to the tree, from which a trajectory 
simulated for a fixed number of steps or until a terminal state is reached. Such trajectories are generated using a
rollout policy; which is typically fast to compute---for instance random and
uniform.  The outcome of this trajectory---
i.e., cumulative discounted rewards along the trajectory---is used to update the value estimates of 
the nodes in the tree that lie along the path from the root to the expanded node.}

\begin{figure}[h]
\centering
\includegraphics[width=0.8\textwidth]{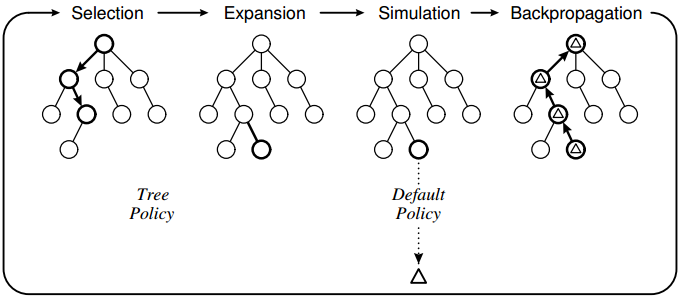}
\caption{Schematic description of MCTS. MCTS methods four components: (I) Selecting a node which has a successor not included in the tree. This node is also referred to as the leaf. Selection of a leaf is determined by a tree policy. (II) A successor of the chosen leaf is added to the tree. (III) A rollout (i.e., trajectory) is sampled using a rollout policy. (IV) Cumulative discounted rewards obtained from the rollout is propagated and backed-up up the tree. Reprinted with permission from \cite{browne2012}. } 
\label{main}
\end{figure}

The goal of MCTS is to select the action at the root with the highest expected
value (EV), that is $Q^*$. This is usually achieved via estimating EVs of root
actions from the returns of the samples collected. \edited{The quality of the
chosen action depends primarily on the tree policy rather than the roll-out
policy. UCT \cite{kocsis2006} and its variants are perhaps the most widely used
tree policies.} For UCT, which we introduce formally later, the EV of a root
action is estimated via the sample mean returns of all rollouts performed from
descendant nodes of the root action. The quality of these estimates depends on
how the tree policy distributes computational effort.  \edited{Next, we describe UCT in more
as well as a modified version of it, which leverages computation values for the tree policy
at the root node.}




\subsection{UCT}

Upper Confidence Bounds applied to trees (UCT) \cite{kocsis2006} adapts a
successful multi-armed bandit algorithm called UCB1 to MCTS. More specifically,
UCT's tree policy applies the UCB1 algorithm recursively down the tree starting
from the root node. At each level, UCT selects the most promising child using
\begin{equation} 
\arg \max_{\nu' \textrm{ is a child of } \nu} \hat{Q}(\nu') + c \sqrt{\frac{2 \log N(\nu)}{N(\nu')}}
\end{equation}
where $N(\nu')$ is the number of times the node $\nu'$ is visited, and
$\hat{Q}(\nu')$ is the average rewards obtained by performing a rollout from
$\nu'$ or one of its descendants, and $c$ is a positive constant, which is
typically selected empirically. The second term assigns higher scores to nodes
that are visited less frequently. As such, it can be thought of as an
exploration bonus.

UCT is simple and has successfully been utilized for many applications.
However, it has also been noted \cite{tolpin2012, hay2012} that UCT's goal is
different from that of approximate planning. UCT attempts to ensure a high net
\emph{simulated} worth for the actions that are taken during the Monte Carlo
simulations that comprise planning. However, all that actually matters is the
\emph{real} worth of the single action that is ultimately taken in the world
after all the simulations have terminated. Failing to simulate an action
because of a low simulated net worth might slow down discovery of its superior
or inferior quality. This problem has been partially address by leveraging
information values---which shifts the focus from the quality of imaginary
scenarios to the quality of the immediate real actions---as we explain next.

\subsection{VOI-based tree policy}

\begin{figure}[htbp]
\begin{center}
\includegraphics[width=0.7\textwidth]{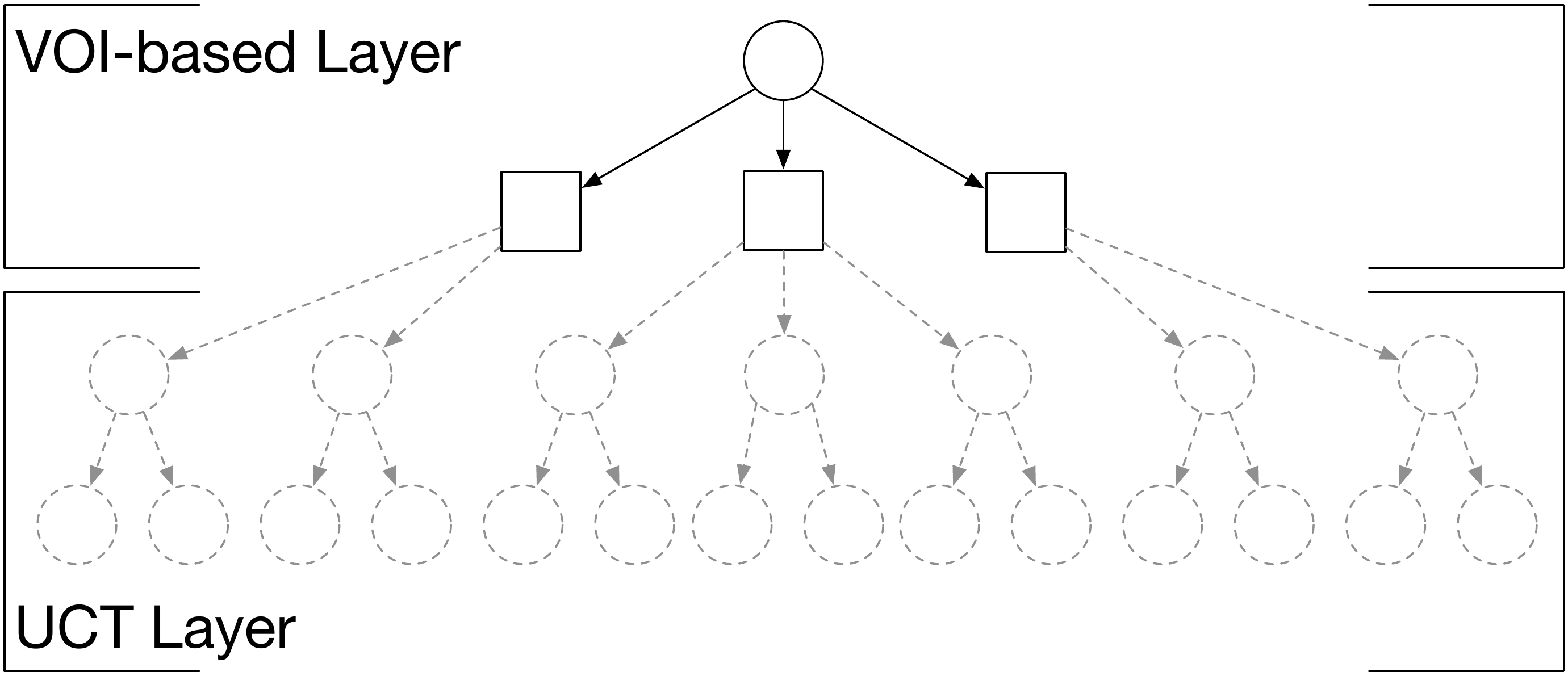}
\caption{VOI-based is a hybrid algorithm, leveraging VOI's for the root actions, and UCT for the rest.}
\label{fig:voi}
\end{center}
\end{figure}

Tolpin and Shimony \cite{tolpin2012} and Hay et. al. \cite{hay2012} tackle the
the aforementioned problem of UCT by suggesting a hybrid sampling scheme in which the
policy for searching the tree employs UCT at actions one step beneath the root,
but calculates value of information (VOI) estimates for the root actions as illustrated in Figure~\ref{fig:voi}. VOIs
quantify the worth of the \emph{information} that is expected to be gained from
a simulation about the ultimate choice, rather than the expected worth of the
simulation itself. These authors demonstrated empirically that even this simple
modification could result in significant performance improvements over UCT.

We can map the VOI definition used in these works to the definitions we have given in Chapter~\ref{ch:voc}. Specifically, VOI in \cite{tolpin2012, hay2012} 
is equivalent to using $1$-step static values (i.e., $\phi_1(s,a | \ot)$) for valuing root actions, and maximizing their respective $\VOCp$'s as defined in Equation~\ref{voc2mdp}. Specifically,

\begin{align}
\textrm{VOI}(\omega; \ot, s) &= \VOCp(\omega; \phi_1, \ot, s) \\
&= \mathbb{E}_{o \sim \omega}\left[\max_{a \in \mathcal{A}_s} \phi_1(s,a | \ot \oplus o) - \phi_1(s, \arg \max_{a \in \mathcal{A}_s}  \phi_1(s,a | \ot) | \ot \oplus o) \right]
\end{align}

Consequently, we can point out three major limitations of VOI's as utilized by these studies, compared to our formulations:
\begin{itemize}
\item \textbf{Evaluating computation values only for the root node of the tree.} Perhaps the most significant shortcoming, as the authors also point out, is that the VOI-based algorithms only compute computation values concerning the root nodes. In other words, computation values are utilized only for choosing the most promising immediate subtree, then UCT is utilized for picking the best node to sample from within the subtree. 

We know that as $n$ increases, $\phi_n(s,a | \ot)$ approaches
$Q^*(s,a)$, independent of $t$. However, VOI-based algorithms value actions according to $\phi_1(s,a | \ot)$ which is likely to be too crude in many cases. $\VOC$-greedy algorithms can work with any $n$. Thus, they offer not only better value representations, but also the ability to select computations in a finer granularity.

\item \textbf{Not accounting for distal policy improvement.}
As discussed in Chapter~\ref{ch:voc}, $\VOCp$ underestimates the gain due to a computation---as it is only concerned with the policy improvement at the root state---ignoring the effect of policy improvement in other states. Note that this is orthogonal to the point raised above as $\VOCp(\omega; \phi_n, \ot, s) \le \VOC(\omega; \phi_n, \ot, s)$ for $n > 1$ as well.

\item \textbf{Not accounting for optionality.} 
As we have discussed in Chapter~\ref{ch:voc}, the cardinal difference between $\psi_n$ and $\phi_n$ is that the former also takes the uncertainties in action values into account, whereas the latter only accounts for the expectations. 

From an economics perspective, the difference between the dynamic and static values (i.e., $\psi_n - \phi_n$) can be considered as the value of optionality---as it reflects how much one can possibly gain by changing one's mind in the light of new information. In other words, $\VOC(\psi_n)$-greedy pays a cost in terms of
potentially choosing the action with lower expected value given what is
known \textit{now}, but maintains the flexibility of allowing itself to
find greater values in the future by the time the agent will have
changed state and will have sampled further.

\end{itemize}

\section{The Value of Computation in MCTS}
In the previous chapter, we have shown that $\VOC(\phi_n)$ can be calculated with reasonable efficiency, and gave an approximation for $\VOC(\psi_n)$, which is otherwise very costly to compute. We now give a more complete and concrete picture by describing how $\VOC(\phi_n/\psi_n)$-greedy can be used for MCTS.

As in the VOI-based algorithm, we have hybrid scheme, where we utilize
UCT as a subroutine to obtain information about states that lie beyond
the spatial horizon. As illustrated in
Algorithm~\ref{alg:voc-greedy}, we keep track of the state of UCT trees
for previously sampled frontier nodes. That is, each time a frontier
node is sampled, UCT subroutine will be invoked, which will in turn
expand the tree by a single node, from which a roll-out will be
performed. We leverage UCT for sampling frontier nodes, rather than
directly performing a roll-out from the frontier node itself, as the
value estimated by UCT will converge to the optimal value (i.e., $Q^*$)
asymptotically. 
\edited{The reason for utilizing $\VOC$'s for the top of the tree/graph---meaning closer to the root/sink---and UCT policy for the bottom part is that the tree policy on the bottom has a smaller impact compared to the one at the top due to discounting among other reasons. Therefore, it is more efficient to save the relatively expensive $\VOC$ computations for the top nodes, and utilize the UCT policy---which can be cheaply evaluated---for the rest.}

\vspace{1cm}
\begin{algorithm}[H]
\SetAlgoLined
\KwIn{Current state $s$}
\KwOut{Selected action $\alpha$ to perform}
Create a partial search graph/tree by expanding state $s$ for $n$
steps \label{alg:tree}\;
Initialize the computation set $\Omega = \{ \langle s', a' \rangle : \nsr \textrm{ from }s\}$ \;
Initialize a partial function $U$, that maps states to UCT trees \;
$t \leftarrow 0$; $o_{1:t} \leftarrow []$  \tcc*[r]{$o_{1:t}$ is an empty vector} 
Initialize $\phi_n/\psi_n$ \;
 \Repeat{computation budget is reached (OR  $\max_{\omega \in \Omega} \textrm{VOC}(\omega; \phi_n/\psi_n, \ot, s) < C$)}{
  $\langle s^*, a^* \rangle = \arg \max_{\omega \in \Omega} \VOC(\omega; \phi_n/\psi_n, \ot, s)$ \label{alg:comp_pol}\;
  $s^\dagger \sim T(s^*, a^*, \cdot)$ \;
  \If{$U(s^\dagger)$ is not defined}{
  Initialize a UCT-tree rooted at $s^\dagger$ \;
  Define $U(s^\dagger)$, which maps to the UCT-tree from the previous step \;
}
Expand $U(s^\dagger)$ and perform a roll-out once \; 
Store the roll-out return in $o$, together with the $\langle s^*, a^* \rangle$ information \;
$o_{1:{t+1}} \leftarrow \ot \oplus o$; \: $t \leftarrow t + 1$ \;
Update $\phi_n/\psi_n$ using $\ot$ \label{alg:update}\;
}
$\alpha = \arg \max_{a \in \As} \phi_n/\psi_n(s,a | \ot)$ \label{alg:pol} \; 
\caption{$\VOC(\phi_n/\psi_n)$-greedy for MCTS}
\label{alg:voc-greedy}
 \end{algorithm}
\vspace{1cm}

Similarly, UEB can also easily be adapted to MCTS. In Algorithm~\ref{alg:voc-greedy}, we need to replace Line~\ref{alg:comp_pol} with $\arg \min_{\langle s', a' \rangle \in \Omega} \partial \lsalpha / \partial n^{s'a'}$ where $\alpha = \arg \max_{a \in \As} \lsa $. Then, when selecting the final action to take, Line~\ref{alg:pol} should be replaced with $\alpha = \arg \max_{a \in \As} \lsa $ as well.

Incorporation of new information is accomplished in Line~\ref{alg:update} by for example using independent, or dependent, conjugate Normal priors.

\edited{
Note that instead of uniformly expanding a search graph as done in Line~\ref{alg:tree}, one can, in principle, incrementally and asymmetrically build a tree as done in MCTS. In this case, at a given time there will be a number of frontier state-actions which we can expand. Therefore, we could compute the static and dynamics values of root actions by trickling the values of these frontier nodes up to the root. However, the problem is, all the previous computations would have been performed upon the nodes that lie between the root and the frontier. Therefore, one needs further assumptions to be able to relate the information about the intermediary nodes to the frontier nodes. One possible way to relate values of state-actions of different depths is to treat the depth as a \emph{context} of a bandit problem and use GPs as in \cite{krause2012}.
}

\chapter{Experiments}
\begin{chapquote}{Ennius (quoted by Cicero), \textit{Medea}}
``If the wise man cannot benefit himself, his wisdom is vain.''
\end{chapquote}

In the previous section, we discuss a $\VOC$-greedy policy can be utilized in
an MCTS setting. Here we benchmark $\VOC$-greedy policies, implemented as MCTS
tree policies, against some of the most popular and/or relevant policies,
namely UCT \cite{kocsis2006}, VOI-based \cite{hay2012}, Bayes UCT
\cite{tesauro2010}, and Thompson sampling for MCTS \cite{bai2013} in two
different environments. We first introduce the latter two of the aforementioned
policies. Then, we move on to the experiment environments, and their respective
results.

\section{Benchmark policies}

Among the four MCTS policies we evaluate against $\VOC$-greedy algorithms, we
have already discussed UCT and VOI-based in Section~\ref{sec:mcts}. Now we
briefly discuss Bayes UCT and Thompson sampling.

What we refer to as Bayes UCT is a Bayesian version of UCT by Tesauro et. al.
\cite{tesauro2010}. Bayes UCT assumes normally distributed action values, and
estimates their mean and scale parameters using conjugate Normal priors. Given
the posterior estimations, the tree policy uses a UCT-like scoring which is a
function of the posterior mean and scale parameters. Like $\VOC$-greedy
policies, Bayes UCT too assumes a fixed spatial horizon, and propagates the
estimated action values up the tree. Therefore, its computational cost is
similar to that of $\VOC(\phi_n)$-greedy. In our implementation, we use UCT to
obtain samples for nodes that lie beyond the spatial horizon as we also do for
$\VOC(\phi_n)$-greedy.

Bai et. al. \cite{bai2013} adapts Thompson sampling---a widely used, optimal
bandit policy---to MCTS. In the original publication, authors use conjugate
Dirichlet and Normal-Gamma distributions to estimate action values. However, in
order to conform with our implementations of Bayes UCT and $\VOC$-greedy, we
use conjugate Normal priors with known variance instead of using Normal-Gamma.
The tree policy uses Thompson sampling recursively down the tree by drawing
action value samples from the posterior distributions and choosing the action
with the highest value sample.

For the experiments we explain next, we select the hyperparameters of all the
policies using grid search.

\section{Bandit-trees}

\begin{figure}[h]
\begin{center}
\includegraphics[width=0.7\textwidth]{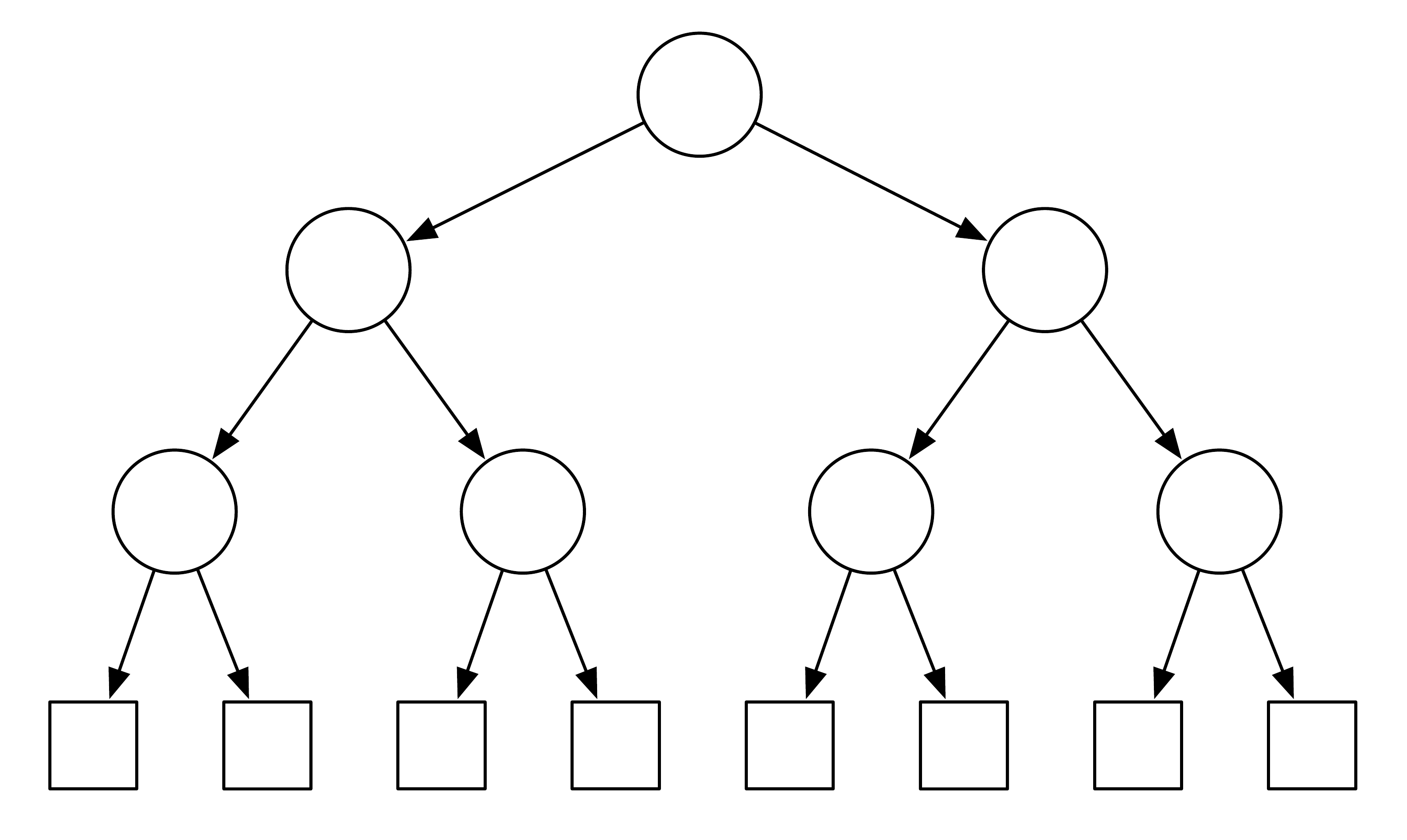}
\caption{A bandit tree of depth $3$. Circles denote states, and squares denote bandit arms.}
\end{center}
\label{bandit-tree}
\end{figure}

The first environment in which we evaluate the aforementioned MCTS methods is
an MDP composed of a complete binary tree of a varying depth $d$. The leaves of
the tree are the bandit arms, whose expected rewards are sampled from a
multivariate Gaussian distribution. The non-leaf nodes are the states. At the
each state, the agent needs to choose between \texttt{LEFT} and \texttt{RIGHT}
actions. \texttt{LEFT} transitions to the left subtree (of depth $d-1$) with
probability $p$ and to the right subtree with probability $1-p$. The same holds
for \texttt{RIGHT}, but with swapped transition probabilities. In
Figure~\ref{bandit-tree}, we illustrate a bandit tree of depth $3$.

We allocate a fixed computational budget to all methods. Each computation
allows the actor to sample a single bandit arm, which is determined by the tree
policy of the agent. After using up the budget, the agent needs to take an
action. In this experiment, we are concerned with the quality of the very
first action (taken at the root node), as quantified by the simple regret
for computation time $t$ as,
\begin{equation}
\SR(\texttt{ROOT}, \pi_t) \coloneqq \max_{a \in \{\texttt{LEFT}, \texttt{RIGHT} \}} Q^*(\texttt{ROOT}, a) - Q^*(\texttt{ROOT}, \pi_t(\texttt{ROOT})) \: ,
\end{equation}
assuming a deterministic policy $\pi_t: \Sc \rightarrow \A$---as all the
aforementioned MCTS policies are deterministic\footnote{Note to be confused by
meta-policies such tree policies or rollout policies, which can be
non-deterministic.}. Note that $\pi$ is subscripted by $t$, as the policy
depends on the past computations.

There are different dimensions along which we can vary the environment, such as
the depth of trees, and transition probabilities. However, our simulations show
that the distribution from which the expected bandit rewards are sampled is
perhaps the most detrimental---especially the degree of correlation among the
arms. Therefore, we present the results of two conditions here, one in which
the expected bandit arm rewards are correlated, and one in which they are not.

For the correlated case, we obtain the exact form of correlation as the
following. We assume that the bandit arms lie uniformly spaced on a line with
unit spacing. For example, for trees of depth $7$, where there are $128$
bandits whose respective coordinates are $0, 1, \dots, 127$. Then, we obtain
the covariance matrix $\Sigma$ using a kernel function, where the distances
among the bandits can be computed from their coordinates. The expected bandit
rewards are sampled from $\mathcal{N}(0.5, \Sigma)$ i.i.d. for each environment
instance. The agent cannot observe the expected rewards directly, but can have
noisy samples of them by performing computations, where the noise is drawn from
$\mathcal{N}(0, 0.01)$ i.i.d.. For the uncorrelated case, we use a white-noise
kernel function.

In our implementations of the algorithms, $\VOC$-greedy and Bayes-UCT have a
fixed spatial horizon of $4$. That is, they unroll the current state for $4$
steps, resulting in $2^4=16$ many frontier nodes. We estimate the posterior
values of these frontier nodes by assuming a conjugate multivariate prior over
the set of frontier nodes for both methods.

\begin{figure}[htbp]
\begin{center}
\includegraphics[width=0.7\textwidth]{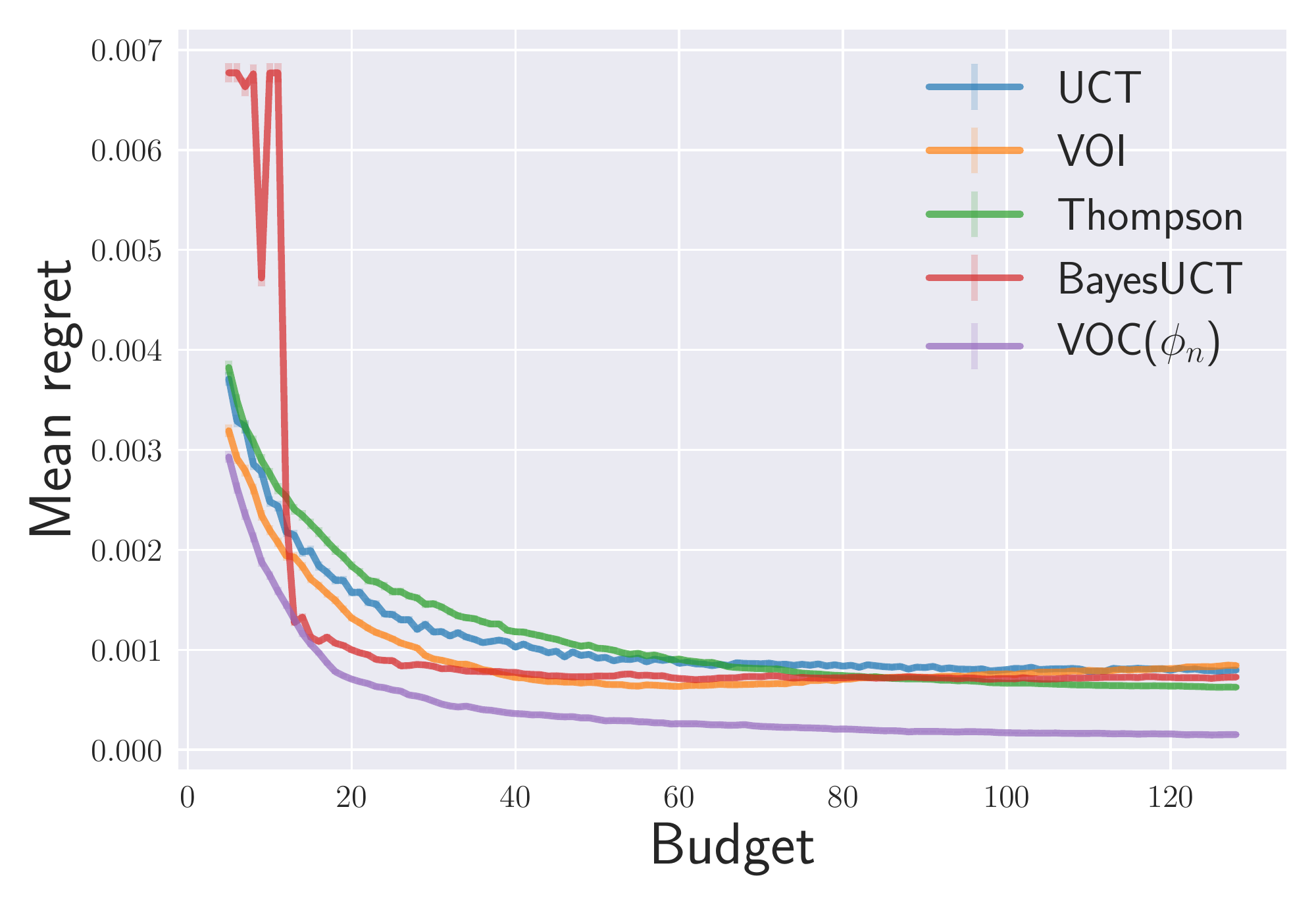}
\label{fig:cor-bt}
\caption{Mean regret (over $10$k environment samples) as a function of computation budget for bandit-trees with \emph{correlated} expected bandit rewards. The plots include mean squared error bars, which are too small to be visible.}
\end{center}
\end{figure}

\begin{figure}
\hfill
\subfigure[Mean regret as a function of computation budget]{\includegraphics[width=0.48\textwidth]{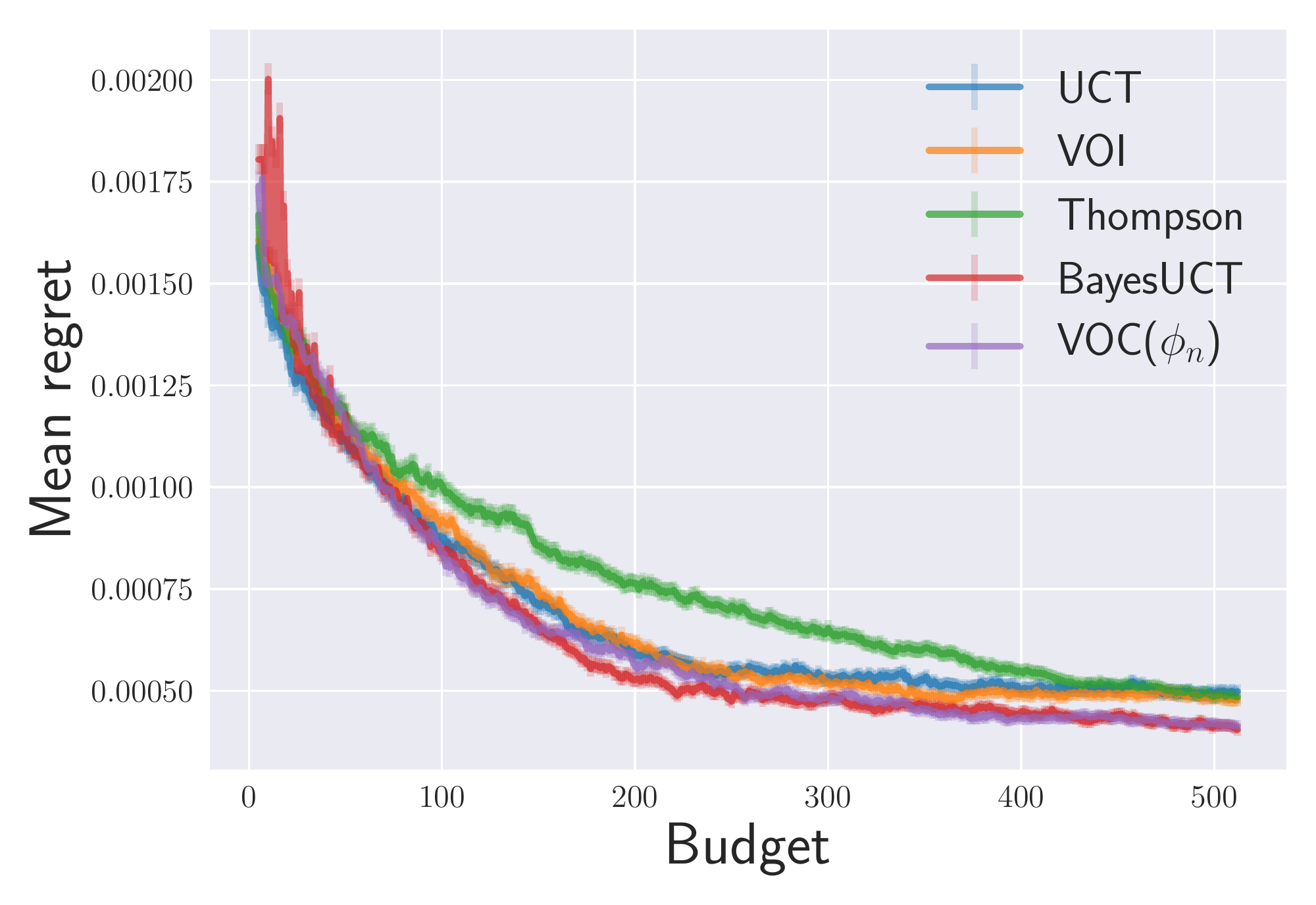}}
\hfill
\subfigure[Cumulative mean regret as a function of computation budget]{\includegraphics[width=0.48\textwidth]{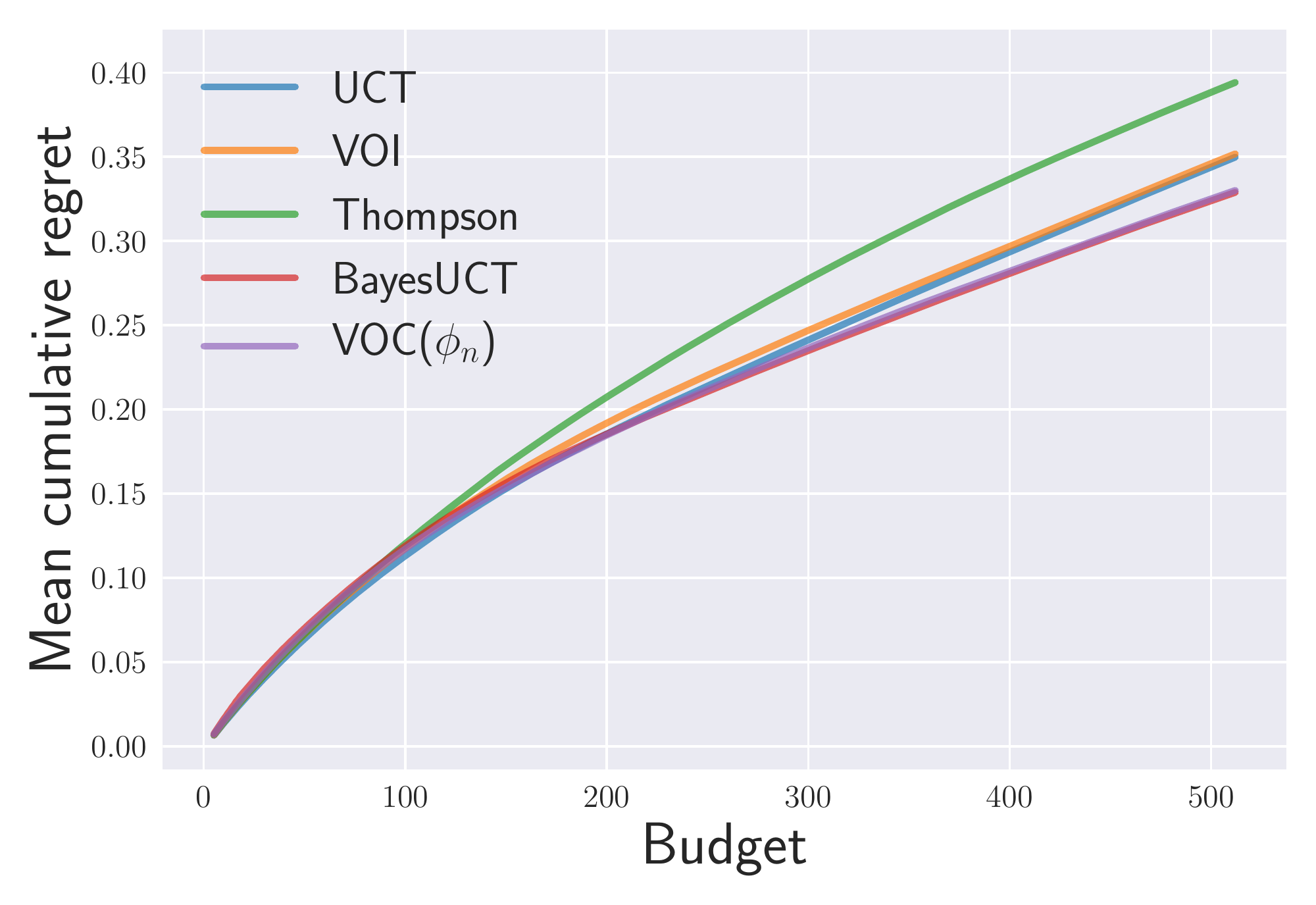}}
\hfill
\caption{Mean regret (over $5$k environment samples) as a function of computation budget for bandit-trees with \emph{uncorrelated} expected bandit rewards. Panel a includes mean squared error bars.}
\label{fig:ind-bt}
\end{figure}

Figure~\ref{fig:cor-bt} shows the results in the case with correlated bandit
arms. We see that $\VOC(\phi_n)$-greedy overperforms all other methods.
However, this might be because $\VOC(\phi_n)$-greedy can quickly learn and
exploit the correlations. In order to control for this, we can also look at a
case where the bandit expected rewards are uncorrelated, or equivalently,
sampled from a white noise kernel. Results of which are presented in
Figure~\ref{fig:ind-bt}. As we can see $\VOC(\phi_n)$-greedy and Bayes UCT
performs equally well, and better than the other policies. This implies that
the good performance of $\VOC(\phi_n)$-greedy is not solely explotaiton of the
correlation structure.

\section{Peg solitaire}

\begin{figure}[htbp]
\begin{center}
\includegraphics[width=0.8\textwidth]{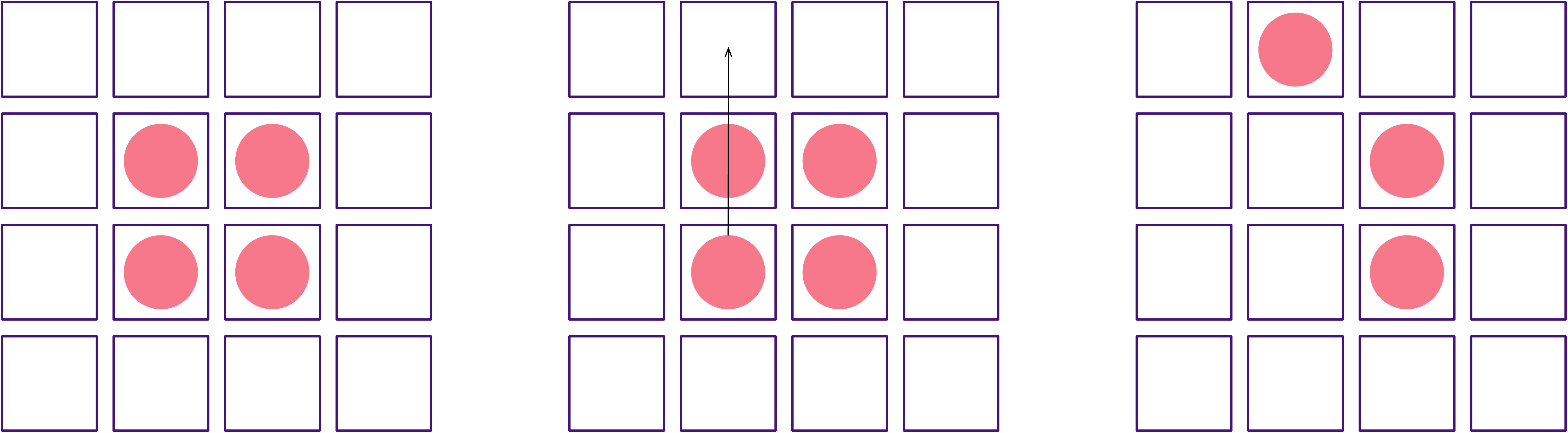}
\caption{Illustration of how pegs move. The bottom left peg is moved up, by skipping over its neighbor---which is then removed from the board.}
\label{ps_example}
\end{center}
\end{figure}

Peg solitaire---also known as Solitaire, Solo, or Solo Noble---is a
single-player board game. The game takes place on a board, where some holes are
occupied by pegs. A valid move is to move a peg over a neighboring peg into a
new unoccupied position two positions away, and then to remove the neighboring
peg as illustrated in Figure~\ref{ps_example}. The objective, for our purposes,
is to minimize the number of pegs remaining on the board.

\begin{figure}[htbp]
\begin{center}
\includegraphics[width=0.7\textwidth]{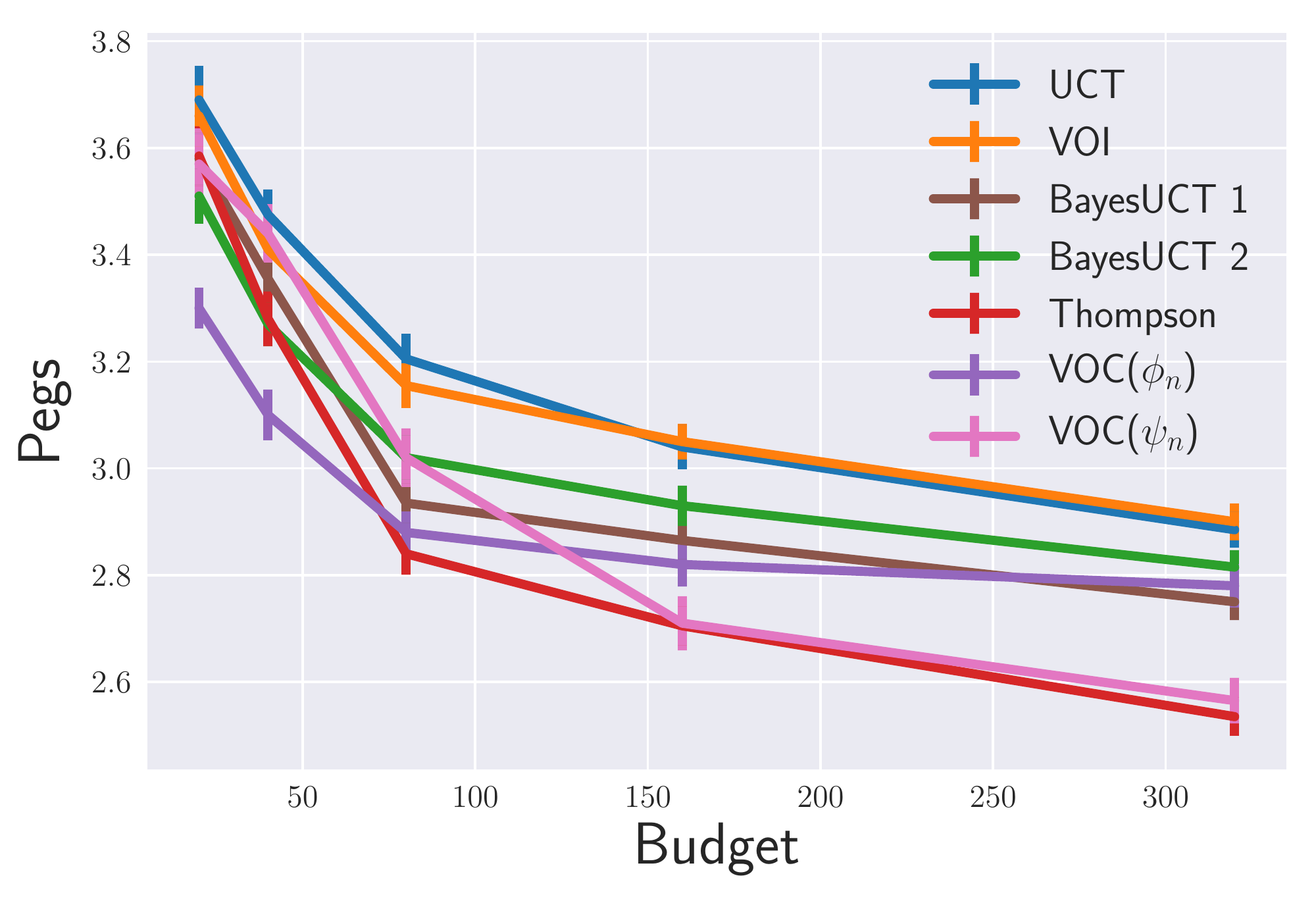}
\caption{The average number of pegs remaining on the board as a function of the computation budget. The bars denote the mean squared errors.}
\label{ps}
\end{center}
\end{figure}

We use a smaller board than the standard game because, especially in mid-game
branching factor is rather high\footnote{For example four pegs arranged a
square form results in eight possible moves. Though, typically branching factor
is in the order of number of pegs on the board.}. More specifically, for the
results we present here, we use a $4 \times 4$ board, with $9$ pegs randomly
placed. The results are illustrated in Figure~\ref{ps}, which are averaged over
$200$ initial board position samples.

We can see that $\VOC(\phi_n)$-greedy performs the best for small budget
ranges, which is in line with our intuition as $\phi_n$ is a more accurate
valuation of action values for small computation budgets. For large budgets, we
see that $\VOC(\psi_n)$-greedy performs as well as Thompson sampling, and
better than the rest.

The performance of $\VOC$-greedy algorithms is perhaps not as good in peg
solitaire compared to the previous environment. However, we should also point
out that, our implementations of $\VOC$-greedy algorithms here assume that
action values are independent, even though they are not in reality. Learning of
the correlations is especially important for environments with high branching.
Consider having a search tree of depth $2$ with branching factor of $20$. Then
there are $20^2 = 400$ leaves. In order to have a reasonable estimate of the
leaf values, each leafs needs to be sampled multiple times, which requires a
few thousand samples. However, the same could be done with way fewer samples if
the correlations can be estimated reasonably well.

\chapter{Discussion}
\begin{chapquote}{Seneca, \textit{De Vita Beata}}
``I am satisfied if each day I make some reduction in the number of vices and find fault with my mistakes.''
\end{chapquote}

We concentrated our efforts in addressing the question of how one should value
computations when planning. We provided an exact answer to this question for a
simple class of MDPs\footnote{Where the agent ends up in a terminal state after
$n$ actions, and receives a final reward, the value of which can be estimated
by performing computations.}. Based on this, we proposed planning algorithms
for approximately solving the general class of MDPs, and provided performance
guarantees. In the experiments section, we demonstrated that our algorithms
perform as well as the best of the state-of-the-art in many instances. Below we
discuss certain areas in which our solutions fall short, and improvements are
possible.

\section{Possible areas of improvement}

We divide shortcomings of our approaches and possible improvements we see into
two sections: theoretical, which mostly pertains to how we define static and
dynamic values, and the value of computation; and practical, which mostly
concerns practical implementations of $\VOC$-greedy, especially for MCTS.

\subsection{Theoretical}

\subsubsection{Filling in between static and dynamic values}

The algorithms we propose either use static or dynamic values, which
respectively assume zero or infinitely many future computations. If the number
of future computations is known, our formulation in principle allows for
valuing the root actions accordingly---that is, using a value that falls
between static and dynamic values, reflecting the budget. However, to do that,
one must somehow determine how those computation would be distributed among the
frontier nodes, for which we have no principled answers.

If this problem could be addressed, then the next question would be how many
computations should one predict performing before taking an action? This, in
turn, would allow forming even more accurate value estimates. For example,
given a problem, if the agent correctly predicts performing $10$
computations\footnote{One could also be Bayesian about predicting the future
number of computations, as well as about the previous open problem.} before
taking an action, then the solution of the problem stated above would give the
optimal way to value actions.

\subsubsection{Sparser trees and sampling}

For computing static and dynamic values, we assume a fixed spatial horizon of $n$. However, this severely limits the practical applications of our theory, as performing full Bellman-backups for $n$-steps can be very costly, especially in environments with high branching factors. This could be remedied by incorporating sparse sampling, and growing a sparse tree adaptively, rather than using full backups for a fixed spatial horizon.

This, however, is a theoretically challenging problem. Under our assumptions, it is easy to quantify the impact of an additional sample on the value of an action. However, quantifying the impact of expanding a node, which we think is necessary, is much harder.

\subsubsection{$\VOC$ for offline updates}

There are many reinforcement learning and planning algorithms that sample a set
random transitions from past experiences (or from a model), and apply
Q-learning-like updates. Examples include \emph{Dyna} \cite{sutton1990} and
\emph{prioritized sweeping} \cite{moore1993}. Our current proposal can only
select computations concerning actions that are directly related to the current
state, which we do via $n$-step values. Being able to do the same for arbitrary
state-actions would yield a more principled version of prioritized sweeping.

One possible way of achieving this is through successor representations (SRs) \cite{dayan1993} as recently done by Matter and Daw \cite{mattar2018}---where the authors suggest that (human) hippocampal replay selects offline updates to maximize a $\VOC$-like measure.
However, the SR-based measure will be rather inflexible and underestimate computation values, given SRs are based on fixed policies\footnote{If SRs are computed from a model directly.} or on a
combination of past policies\footnote{If SRs are computed via a temporal-difference mechanism.}, and as such, are incapable of accounting for possible future policies unlike $\VOC(\psi_n)$-greedy.

\subsubsection{Stopping criterion}

For the most part, we conducted our analysis as well as the experiments using a
fixed computation budget. This is in fact the standard in MCTS applications,
where computation budgets are determined a priori using heuristics.

An important advantage of calculating $\VOC$s is that it naturally lends itself
to adaptive stopping criteria \cite{hay2012}. A simple use is that a
computation is performed if and only if the $\VOC$ is higher than a threshold
as in Equation~\ref{eq:stop}. How to set this threshold is an open question.


 
\subsection{Practical}

\subsubsection{Learning correlations}
As we discuss in the experiments sections, being able to learn the correlations of frontier action values is detrimental for performing well. 
We achieve this by obtaining a assuming prior covariance (given by the kernel), and then performing Bayesian updates. However, there are many scenarios where finding a good kernel might require some ``feature engineering''. This is a disadvantage of our $\VOC$-greedy implementations, as it does not necessarily function well ``off the shelf''.

\subsubsection{Efficient computing of $\phi$ and $\psi$ in stochastic environments}
We provide two ways of handling stochastic transitions for computing $\phi$ and $\psi$ (see Section~\ref{sec:stoc}): (I) numerically computing the exact solution; (II) sampling deterministic transitions based on the transition probabilities, and taking the expectation---which is not exact but forms an upper bound, but at a cheaper cost. Given the ubiquity of stochastic environments, having better algorithms here would have great practical value.

\subsubsection{Meta-computational costs}

In the experiments section, we show that our $\VOC$-greedy algorithms perform
better than most of the state-of-the-art given the same number of computations
(i.e., rollouts). However, there are significant difference in the number
meta-computations---that is, computations used for selecting the rollout-level
computations---these policies perform. Especially, computing the
$\VOC(\psi)$-greedy policy is very costly. $\VOC(\phi)$-greedy has a comparable
meta-computational cost to that of Bayes UCT and Thompson sampling, but only if
the transitions are deterministic.

\section{Final thoughts}

We provide a normative theory of how computations should be valued in a
planning setting. Our work overcomes some of the significant limitations of
existing $\VOC$-like measures for planning by extending computation values to
non-immediate actions and by accounting for the impact of possible future
computations. Even though these improvements come at a large
(meta-)computational cost, limiting its immediate practical applications, we
hope that by defining what the value of a computations is more accurately, we
can understand planing better, and thus plan better.

\clearpage

\backmatter



\bibliographystyle{unsrt}
\bibliography{thesis} 

\end{document}